\documentclass[10pt,journal,compsoc]{IEEEtran}
\usepackage{CJK} 

\usepackage{ifpdf}
\ifpdf
\else
\fi
\ifCLASSINFOpdf
   \usepackage[pdftex]{graphicx}
   \graphicspath{{../pdf/}{../jpeg/}}
   \DeclareGraphicsExtensions{.pdf,.jpeg,.png}
\else
   \usepackage[dvips]{graphicx}
   \graphicspath{{../eps/}}
   \DeclareGraphicsExtensions{.eps}
\fi
\usepackage{amsmath}

\usepackage{pifont} 

\usepackage{array}

\ifCLASSOPTIONcompsoc
 \usepackage[caption=false,font=footnotesize,labelfont=sf,textfont=sf]{subfig}
\else
 \usepackage[caption=false,font=footnotesize]{subfig}
\fi

\usepackage{stfloats}
\ifCLASSOPTIONcaptionsoff
 \usepackage[nomarkers]{endfloat}
\let\MYoriglatexcaption\caption
\renewcommand{\caption}[2][\relax]{\MYoriglatexcaption[#2]{#2}}
\fi
\let\MYorigsubfloat\subfloat
\renewcommand{\subfloat}[2][\relax]{\MYorigsubfloat[]{#2}}
\usepackage{xspace}
\makeatletter
\DeclareRobustCommand\onedot{\futurelet\@let@token\@onedot}
\def\@onedot{\ifx\@let@token.\else.\null\fi\xspace}

\makeatother

\usepackage{enumitem}
\usepackage{caption}
\usepackage{times}
\usepackage{epsfig}
\usepackage{graphicx}
\usepackage{amsmath}
\usepackage{amssymb}
\usepackage{eso-pic}
\usepackage{helvet}
\usepackage{courier}
\usepackage{url}
\usepackage{mathrsfs}
\usepackage{multirow}
\usepackage{comment}
\usepackage{marvosym}
\usepackage{bbm}
\usepackage{enumerate}
\usepackage{soul}
\usepackage[utf8]{inputenc}
\usepackage{amsfonts}
\usepackage{nicefrac}
\usepackage{microtype}
\usepackage{booktabs} 
\usepackage{algorithm}
\usepackage{algorithmicx}
\usepackage{algpseudocode}
\usepackage{makecell}
\usepackage{xcolor}
\usepackage{multirow}
\usepackage{comment}

\usepackage{bbding}
\usepackage[numbers]{natbib}
\usepackage[breaklinks=true,bookmarks=false,colorlinks]{hyperref}
\usepackage{cleveref}
\usepackage[normalem]{ulem}
 
\usepackage{booktabs,tabularx,array,makecell}
\newcolumntype{P}[1]{>{\raggedright\arraybackslash}p{#1}} 
\newcolumntype{Y}{>{\raggedright\arraybackslash}X}         

\usepackage{color, colortbl}
\usepackage{xcolor}

\definecolor{Gray}{gray}{0.9}

\definecolor{roadcolor}{RGB}{234,51,246}
\definecolor{sidewalkcolor}{RGB}{68,8,72}
\definecolor{parkingcolor}{RGB}{241,156,249}
\definecolor{othergroundcolor}{RGB}{160,32,76}
\definecolor{buildingcolor}{RGB}{246,202,69}
\definecolor{carcolor}{RGB}{111,149,238}
\definecolor{truckcolor}{RGB}{74,32,172}
\definecolor{bicyclecolor}{RGB}{136,227,242}
\definecolor{motorcyclecolor}{RGB}{37,59,146}
\definecolor{othervehiclecolor}{RGB}{96,81,242}
\definecolor{vegetationcolor}{RGB}{79, 173, 50}
\definecolor{trunkcolor}{RGB}{126, 65, 22}
\definecolor{terraincolor}{RGB}{171, 238, 105}
\definecolor{personcolor}{RGB}{234, 60, 49}
\definecolor{bicyclistcolor}{RGB}{234, 66, 195}
\definecolor{motorcyclistcolor}{RGB}{138, 42, 90}
\definecolor{fencecolor}{RGB}{238, 128, 69}
\definecolor{polecolor}{RGB}{252, 241, 161}
\definecolor{trafficsigncolor}{RGB}{233, 51, 35}
\definecolor{color1}{RGB}{176, 36, 24}
\definecolor{color2}{RGB}{119,185,0}
\definecolor{color3}{RGB}{0, 0, 200}
\definecolor{colorofteaser}{RGB}{176, 36, 24}
\definecolor{LightGrey}{rgb}{.9,.9,.9}
\definecolor{White}{rgb}{1.,0.,1.}
\definecolor{first}{rgb}{.8,.0,.0}
\definecolor{second}{rgb}{.0,.6,.0}
\definecolor{third}{rgb}{.0,.0,.8}
\definecolor{ceiling}{RGB}{214,  38, 40}   %
\definecolor{floor}{RGB}{43, 160, 4}     %
\definecolor{wall}{RGB}{158, 216, 229}  %
\definecolor{window}{RGB}{114, 158, 206}  %
\definecolor{chair}{RGB}{204, 204, 91}   %
\definecolor{bed}{RGB}{255, 186, 119}  %
\definecolor{sofa}{RGB}{147, 102, 188}  %
\definecolor{table}{RGB}{30, 119, 181}   %
\definecolor{tvs}{RGB}{160, 188, 33}   %
\definecolor{furniture}{RGB}{255, 127, 12}  %
\definecolor{objects}{RGB}{196, 175, 214} %
\definecolor{car}{rgb}{0.39215686, 0.58823529, 0.96078431}
\definecolor{bicycle}{rgb}{0.39215686, 0.90196078, 0.96078431}
\definecolor{motorcycle}{rgb}{0.11764706, 0.23529412, 0.58823529}
\definecolor{truck}{rgb}{0.31372549, 0.11764706, 0.70588235}
\definecolor{other-vehicle}{rgb}{0.39215686, 0.31372549, 0.98039216}
\definecolor{person}{rgb}{1.        , 0.11764706, 0.11764706}
\definecolor{bicyclist}{rgb}{1.        , 0.15686275, 0.78431373}
\definecolor{motorcyclist}{rgb}{0.58823529, 0.11764706, 0.35294118}
\definecolor{road}{rgb}{1.        , 0.        , 1.        }
\definecolor{parking}{rgb}{1.        , 0.58823529, 1.        }
\definecolor{sidewalk}{rgb}{0.29411765, 0.        , 0.29411765}
\definecolor{other-ground}{rgb}{0.68627451, 0.        , 0.29411765}
\definecolor{building}{rgb}{1.        , 0.78431373, 0.        }
\definecolor{fence}{rgb}{1.        , 0.47058824, 0.19607843}
\definecolor{vegetation}{rgb}{0.        , 0.68627451, 0.        }
\definecolor{trunk}{rgb}{0.52941176, 0.23529412, 0.        }
\definecolor{terrain}{rgb}{0.58823529, 0.94117647, 0.31372549}
\definecolor{pole}{rgb}{1.        , 0.94117647, 0.58823529}
\definecolor{traffic-sign}{rgb}{1.        , 0.        , 0.    }

\definecolor{barrier1}{RGB}{112,128,144}
\definecolor{bicycle1}{RGB}{220,20,60}
\definecolor{bus1}{RGB}{255, 127, 80}
\definecolor{car1}{RGB}{255, 158, 0}
\definecolor{const. veh.1}{RGB}{233, 150, 70}
\definecolor{motorcycle1}{RGB}{255,61,99}
\definecolor{pedestrian1}{RGB}{0,0,230}
\definecolor{traffic cone1}{RGB}{47,79,79}
\definecolor{trailer1}{RGB}{255,140,0}
\definecolor{truck1}{RGB}{255,99,71}
\definecolor{drive. suf.1}{RGB}{0,207,191}
\definecolor{other flat1}{RGB}{175,0,75}
\definecolor{sidewalk1}{RGB}{75,0,75}
\definecolor{terrain1}{RGB}{112,180,60}
\definecolor{manmade1}{RGB}{222,184,135}
\definecolor{vegetation1}{RGB}{0,175,0}

\definecolor{nbarrier}{RGB}{255, 120, 50}
\definecolor{nbicycle}{RGB}{255, 192, 203}
\definecolor{nbus}{RGB}{255, 255, 0}
\definecolor{ncar}{RGB}{0, 150, 245}
\definecolor{nconstruct}{RGB}{0, 255, 255}
\definecolor{nmotor}{RGB}{200, 180, 0}
\definecolor{npedestrian}{RGB}{255, 0, 0}
\definecolor{ntraffic}{RGB}{255, 240, 150}
\definecolor{ntrailer}{RGB}{135, 60, 0}
\definecolor{ntruck}{RGB}{160, 32, 240}
\definecolor{ndriveable}{RGB}{255, 0, 255}
\definecolor{nother}{RGB}{139, 137, 137}
\definecolor{nsidewalk}{RGB}{75, 0, 75}
\definecolor{nterrain}{RGB}{150, 240, 80}
\definecolor{nmanmade}{RGB}{213, 213, 213}
\definecolor{nvegetation}{RGB}{0, 175, 0}

\usepackage{xcolor}
\newcommand{\xj}[1]{\textcolor{red}{#1}}

\begin{document}


\title{Compression Tells Intelligence: Visual Coding, Visual Token Technology, and the Unification}

\author{Xin Jin, \textit{Member, IEEE},
\thanks{Xin Jin (corresponding author) is an assistant professor at the Eastern Institute of Technology, Ningbo, China, (e-mail: jinxin@eitech.edu.cn).
} Jinming Liu, Yuntao Wei, Junyan Lin, Zhicheng Wang, \\
Jianguo Huang, Xudong Yang, Yanxiao Liu, Wenjun Zeng, \textit{Fellow, IEEE}
\\Eastern Institute of Technology, Ningbo}











\IEEEtitleabstractindextext{
\begin{abstract}  


``Compression Tells Intelligence'', is supported by research in artificial intelligence, particularly concerning (multimodal) large language models (LLMs/MLLMs), where compression efficiency often correlates with improved model performance and capabilities. For compression, classical visual coding based on traditional information theory has developed over decades, achieving great success with numerous international industrial standards widely applied in multimedia (e.g., image/video) systems. Except that, the recent emergingvisual token technology of generative multi-modal large models also shares a similar fundamental objective like visual coding: maximizing semantic information fidelity during the representation learning while minimizing computational cost. Therefore, this paper provides a comprehensive overview of two dominant technique families first -- Visual Coding and
Vision Token Technology -- then we further unify them from the aspect of optimization, discussing the essence of compression efficiency and model performance trade-off behind. Next, based on the proposed unified formulation bridging visual coding andvisual token technology, we synthesize bidirectional insights of themselves and forecast the next-gen visual codec and token techniques.
Last but not least, we experimentally show a large potential of the task-oriented token developments in the more practical tasks like multimodal LLMs (MLLMs), AI-generated content (AIGC), and embodied AI, as well as shedding light on the future possibility of standardizing a general token technology like the traditional codecs (e.g., H.264/265) with high efficiency for a wide range of intelligent tasks in a unified and effective manner.  



\end{abstract}

}

\IEEEdisplaynontitleabstractindextext
\maketitle

\section{Introduction}
\IEEEPARstart{A} compelling principle is emerging: ``Compression Tells Intelligence"~\cite{huang2024compression}. This perspective holds that the essence of intelligence is the ability to form compact and effective representations of the world by identifying, modeling, and exploiting patterns within data. The recent success of Large Language Models (LLMs)~\cite{bai2025qwen2,li2024llama} provides strong validation for this concept. Their extraordinary capabilities in reasoning, generation, and in-context learning stem directly from their ability to compress vast linguistic data into powerful internal representations. \emph{As a result, compression efficiency has evolved from a simple engineering metric for storage and bandwidth into a fundamental benchmark for a model's depth of understanding and intelligence.}

This core philosophy naturally extends to the visual domain, where it has inspired two distinct, but strongly related, lines of technological development. The first is \textbf{Classical Visual Coding}~\cite{jpeg2000overview,wallace1991jpeg,bross2013hevc,bross2021vvc,balle2018variational,cheng2020learned}. Grounded in information theory, this field has a long history of success, producing international standards from JPEG~\cite{wallace1991jpeg} to  H.265/HEVC~\cite{bross2013hevc}. These technologies excel at minimizing statistical redundancy to achieve the highest possible pixel-level fidelity, and they form the foundation of our modern multimedia ecosystem. The second line of development is the recently emerging \textbf{Visual Token Technology}~\cite{radford2021clip,zhai2023sigmoid,shang2024prumerge,alvar2025divprune}, which has emerged alongside generative AI and Multimodal Large Language Models (MLLMs)~\cite{bai2025qwen2,liu2023visual,chen2024internvl}. Unlike classical coding, the primary goal of visual tokens is not the perfect reconstruction of pixels, but rather the extraction of crucial semantic information for downstream tasks like visual question answering or image generation. Despite their different approaches, both classical coding and vision technology share the same objective: \emph{to find an optimal balance between information fidelity and computational cost.} 


Despite this shared goal, these two technical families have evolved almost entirely independently. They are pursued by different academic communities (Signal Processing vs. Machine Learning), are based on different theoretical principles (Information Theory vs. Representation Learning), and are evaluated by different criteria (e.g., visual quality vs. downstream task accuracy). This divergence extends to the very purpose of compression itself. Classical coding primarily aims to reduce data size for efficient storage and transmission, thus saving \textbf{bandwidth}. In contrast, visual token technology seeks to create a compact sequence of representations to reduce the \textbf{computational} cost of learning processing by large-scale models like Transformers. This separation has triggered a significant gap. Classical codecs, optimized to minimize bit-rate against signal fidelity, offer unparalleled compression efficiency but their representations are not inherently designed for direct use in AI model architectures. Conversely, visual tokens are explicitly designed to produce compact feature sets that reduce computational load and improve model performance, yet they currently lack the theoretical rigor and compression rates of traditional methods. \emph{We argue that bridging this gap is essential. A unified framework would allow to understand the fundamental trade-off between compression efficiency and model performance more deeply, paving the way for the next generation of visual intelligence.}

\begin{figure*}[t]
    \centering
    \includegraphics[width=1\linewidth]{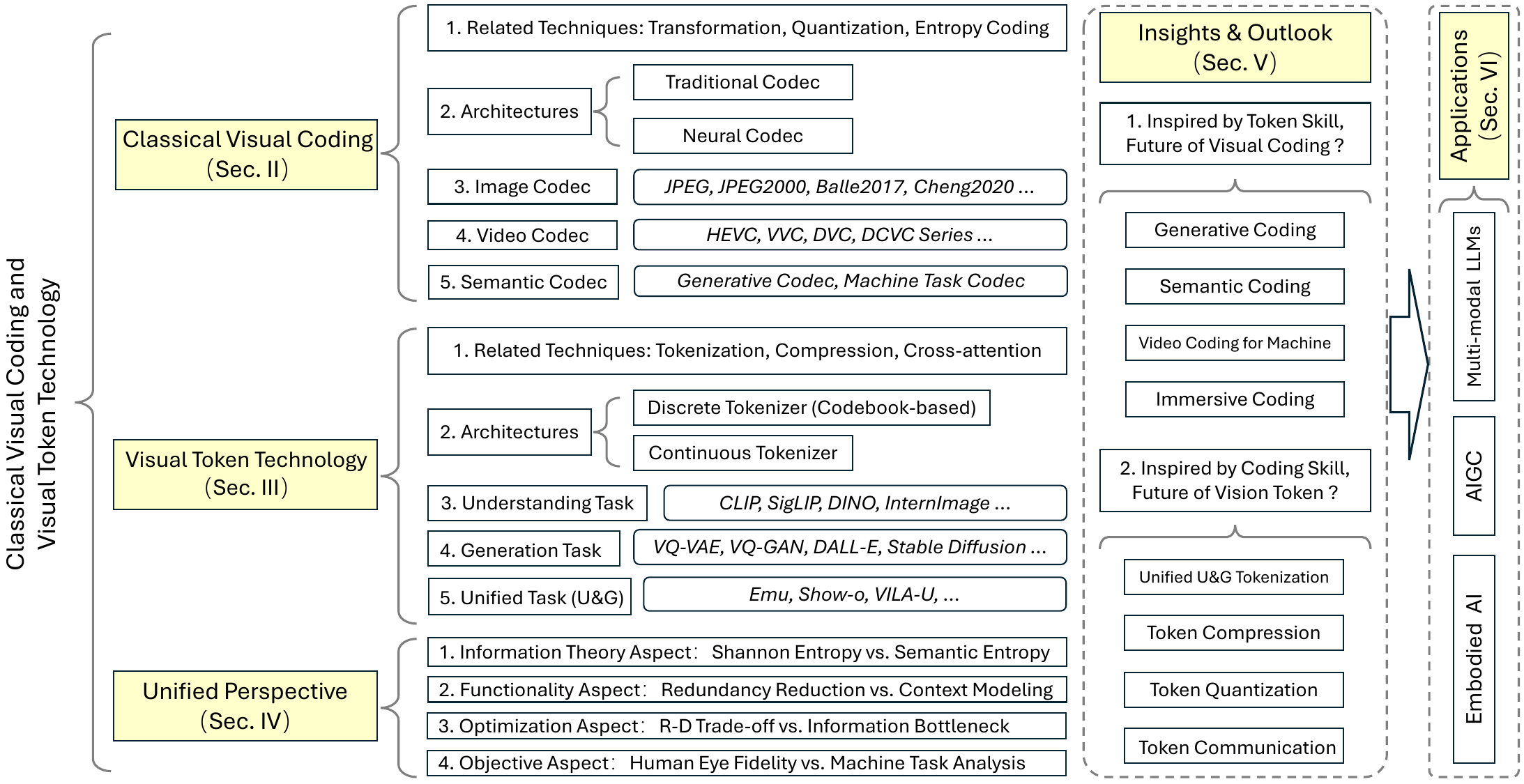}
    \vspace{-4mm}
    \caption{The overall organization of this paper.}
    \vspace{-15pt}
    \label{fig:teaser}
\end{figure*}

As shown in Fig.~\ref{fig:teaser} to bridge this gap and foster innovation between the fields, this paper makes the following key contributions and organizes our paper as follows:
\begin{itemize}
    \item \xj{Section II\&III:} We provide the first systematic review that connects the fields of classical visual coding and emerging visual token technologies, outlining their histories, core principles, and key techniques.
    \item \xj{Section IV\&V:} We propose a theoretical framework that unifies the goals of both visual coding and visual token technology from different perspectives. Based on the unified framework, we distill key insights that allow each field to re-formula and improve the other, forecasting the next-gen visual coding and visual token technology.
    \item \xj{Section VI:} We demonstrate the significant potential of compression technology, particularly focusing on the fast-developing visual token skills rather than well-standardized visual coding, on system-level real-world applications, including MLLMs, AIGC, and Embodied AI.
\end{itemize}

\section{Classical Visual Coding and Codecs}
Classical visual coding~\cite{jpeg2000overview,liu2023learned,minnen2018joint,li2023frequency} seeks to create compact representations of visual data, minimizing the required bits while preserving essential information. This core pursuit facilitates efficient storage and transmission across digital platforms. All realizations share the same three technique primitives: transformation for decorrelation/energy compaction, quantization for discretization and rate control, and entropy coding for lossless compression of syntax symbols. This section provides an overview of classical visual coding, starting with the fundamental techniques, followed by different architectures, specific codec instances for images and videos, and finally, the emerging area of semantic coding.

\subsection{Related Core Techniques}
The foundational principles of nearly all visual coding systems, both traditional and learned, revolve around three core techniques. \textbf{Transformation} is employed to decorrelate the visual data and compact its energy into a smaller set of coefficients~\cite{wallace1991jpeg,liu2023learned,cheng2020learned}. Common transforms include the Discrete Cosine Transform (DCT) in JPEG and many video codecs~\cite{jia2025towards,li2021deep_dcvc}, and more recently, learned non-linear transforms using autoencoders in neural codecs. \textbf{Quantization} is the process of reducing the precision of the transformed coefficients, which is the primary source of lossy compression~\cite{cui2021asymmetric_gainunit,tong2023qvrf}. This step is crucial for controlling the bitrate. \textbf{Entropy coding}~\cite{minnen2020channel,koyuncu2022contextformer,lu2025learned_dcae}, such as Huffman coding or arithmetic coding, is the final stage, where the quantized symbols are losslessly compressed by assigning shorter codes to more probable symbols.

\subsection{Architectures of Visual Coding}


\subsubsection{Traditional Codec}
Traditional codecs, like those famous standards of JPEG, JPEG 2000, HEVC, VVC, etc.,~\cite{jpeg2000overview,jpegai,bross2013hevc,bross2021vvc} are built upon hand-crafted modules that are individually optimized. They typically follow a block-based hybrid coding framework, especially for video. This architecture involves prediction (either spatial for intra-frames or temporal for inter-frames), transformation of the residual, quantization, and entropy coding.

\subsubsection{Neural Codec}
Learned codecs~\cite{liu2023learned,li2023frequency,lu2025learned_dcae}, referred to as neural codecs, replace the hand-crafted components of traditional codecs with deep neural networks. These architectures are trained end-to-end, typically using an autoencoder framework for the transform and learned priors for entropy modeling. This data-driven approach allows for more powerful and adaptive modeling of complex visual data.

\subsection{Image Codec}
\begin{figure}
    \centering
    \includegraphics[width=1\linewidth]{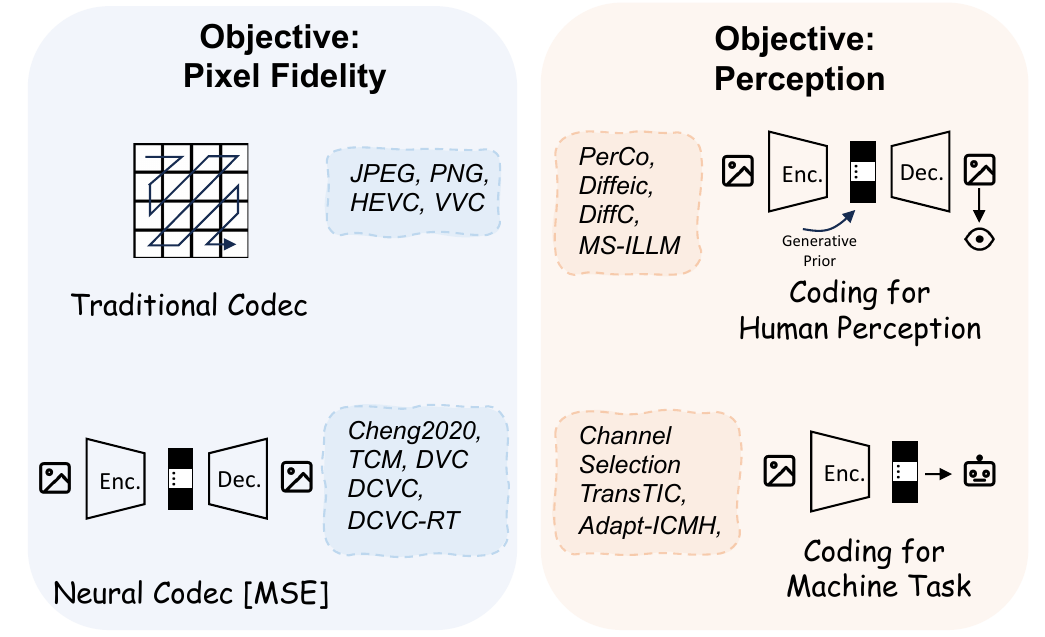}
    \vspace{-15pt}
    \caption{A taxonomy of modern video coding paradigms, categorized by their different optimization objectives. The left branch represents traditional and neural codecs optimized for pixel fidelity (e.g., JPEG~\cite{wallace1991jpeg}, PNG~\cite{boutell1997png}, HEVC~\cite{bross2013hevc}, VVC~\cite{bross2021vvc}, DVC~\cite{lu2019dvc}, DCVC~\cite{li2021deep_dcvc}, etc). The right branch focuses on coding for human perception (e.g., PerCo~\cite{careil2024towards}, Diffeic~\cite{li2024towards}, DiffC~\cite{theis2022lossy}, MS-ILLM~\cite{muckley2023improving}) and coding for machine tasks (e.g., Channel Selection~\cite{liu2022improving}, TransTIC~\cite{chen2023transtic}, Adapt-ICMH~\cite{li2024image}).
}
\vspace{-15pt}
    \label{fig:visual_coding}
\end{figure}


\subsubsection{Traditional Image Codec}
\textbf{JPEG} is the canonical lossy image standard: images are divided into $8{\times}8$ blocks, transformed by a DCT, coefficients are zig-zag scanned, run-length coded, quantized, and entropy-coded via Huffman/arithmetic coding~\cite{wallace1991jpeg}.
\textbf{JPEG2000} improves upon JPEG by using a Discrete Wavelet Transform (DWT), which provides better compression performance and features like scalability and region-of-interest coding.
For lossless coding, \textbf{PNG} applies predictive filtering followed by \texttt{DEFLATE}~\cite{graphics2003specification}. These systems exemplify the classical transform–quantize–entropy pipeline.

\subsubsection{Learned Image Codec}
Learned image codecs retain the same three primitives but replace hand-crafted parts with data-driven ones: an autoencoder provides the transform; (soft) quantization or vector quantization discretizes latents; and a learned prior—commonly a hyperprior and/or an autoregressive or attention-based context model—predicts symbol probabilities for the entropy coder~\cite{balle2018variational,minnen2018joint,cheng2020learned}. Foundational works proposed hyperprior models that leverage side information for more accurate entropy estimation of the latent codes~\cite{balle2018variational,minnen2018joint}. Subsequent research introduced improved network architectures and transforms~\cite{cheng2020learned}. Recent systems (e.g., ELIC and successors) push state-of-the-art rate–distortion, often rivaling or surpassing VVC on standard datasets~\cite{he2022elic,liu2023learned}. These architectures were trained end-to-end to optimize for metrics like PSNR or MS-SSIM. Moreover, even though scalar quantization is known to be suboptimal in the literature, due to the high computational complexity of classical vector quantization schemes in information theory and the difficulty of estimating the rate-distortion function, learned compression algorithms based on neural networks~\cite{yang2023introduction, lei2023neural} have been studied.
Such a learned neural compressor can even approximately recover the optimal vector quantization performance at reasonable complexity~\cite{lei2024approaching}.

\subsection{Video Codec}
\subsubsection{Traditional Video Codec}
Modern block-based hybrids (e.g., \textbf{HEVC/H.265}~\cite{bross2013hevc} and \textbf{VVC/H.266})~\cite{bross2021vvc} extend the image pipeline with motion-compensated prediction (temporal), rich intra prediction (spatial), hierarchical block/tree partitioning (e.g., CTU/QTMT), in-loop filtering (deblocking, SAO), and context-adaptive binary arithmetic coding (CABAC)~\cite{bross2013hevc,bross2021vvc}. Transform choice is typically integerized DCT/DST variants; quantization uses rate–distortion optimized decisions; entropy coding leverages context models tied to local syntax structure. These standards have pushed the rate-distortion (R-D) performance frontier under metrics like PSNR.

\subsubsection{Learned Video Codec}
Learned video codecs (e.g., the \textbf{DCVC series}~\cite{li2021deep_dcvc,jia2025towards}) model motion and residuals directly in the \emph{feature/latent} domain via learned warping/conditioning, with recurrent or GOP-structured inference; the entropy term becomes the cross-entropy of quantized latents under conditional priors, enabling strong rate savings with real-time throughput on GPUs~\cite{jia2025towards}. Similarly, in video, learned approaches have demonstrated superior performance over traditional standards while maintaining practical inference speeds~\cite{jia2025towards}. 

\subsection{Semantic Codec}
Historically, the definition of "essential information" has been tightly coupled with mathematical, pixel-level fidelity. However, the field is undergoing a significant transformation, with optimization objectives evolving from simple pixel-level accuracy to encompass more sophisticated goals tailored for human perception and machine-vision tasks. This has led to the development of semantic codecs, as shown in Fig~\ref{fig:visual_coding}.


\subsubsection{Human-Perception-Oriented Coding}


While pixel-based metrics are computationally convenient, they often correlate imperfectly with subjective quality perceived by human observers. This mismatch motivated a perceptual optimization paradigm, where the objective shifts from minimizing mathematical distortion to maximizing visual realism and appeal.

This paradigm is closely tied to generative models, which can synthesize natural-looking textures and details that pixel-wise losses tend to suppress. GAN-based approaches~\cite{agustsson2019generative} were among the first to demonstrate this advantage, producing reconstructions that are often subjectively preferred over PSNR-optimized counterparts, even when PSNR is lower~\cite{mentzer2020high}. More recently, diffusion models have pushed perceptual compression further, achieving state-of-the-art performance and generating high-fidelity, visually pleasing images even at very low bitrates~\cite{mao2024perceptual,li2024towards,muckley2023improving,theis2022lossy,careil2024towards}. These methods explicitly prioritize plausible synthesis over exact reconstruction, representing a clear departure from pixel-fidelity-oriented optimization.


\subsubsection{Machine-Vision-Oriented Coding}
The most recent and transformative paradigm shift in visual coding is driven by the proliferation of machine-centric applications. In this context, the ultimate consumer of the visual data is not a human, but an AI model performing a task like classification, detection, or segmentation. Consequently, the optimization objective moves away from both pixel fidelity and human perception, focusing instead on preserving the semantic information essential for machine tasks~\cite{chen2023transtic}.

The goal becomes maximizing task accuracy for a given bitrate, leading to a rate-accuracy trade-off. Semantic codecs are designed to identify and allocate more bits to features critical for machine analysis while aggressively compressing irrelevant background information~\cite{li2024misc,liu2022semantic}. Some approaches have demonstrated the benefit of jointly optimizing the compression model and the downstream task network, further improving machine task performance~\cite{li2024human}. To accommodate diverse use cases, scalable bitstreams have been developed that can flexibly serve both human and machine needs from a single compressed representation~\cite{liu2024rate, jin2023semantical}. This evolution has culminated in industry-led standardization initiatives, such as MPEG's Video Coding for Machines (VCM)~\cite{vcm} and JPEG AI~\cite{jpegai}, which aim to create a unified framework that recognizes both perceptual and semantic goals in next-generation codecs.

  \section{VISUAL TOKEN TECHNOLOGY OF MLLMs}
\begin{figure*}[t]
    \centering
    \includegraphics[width=1\linewidth]{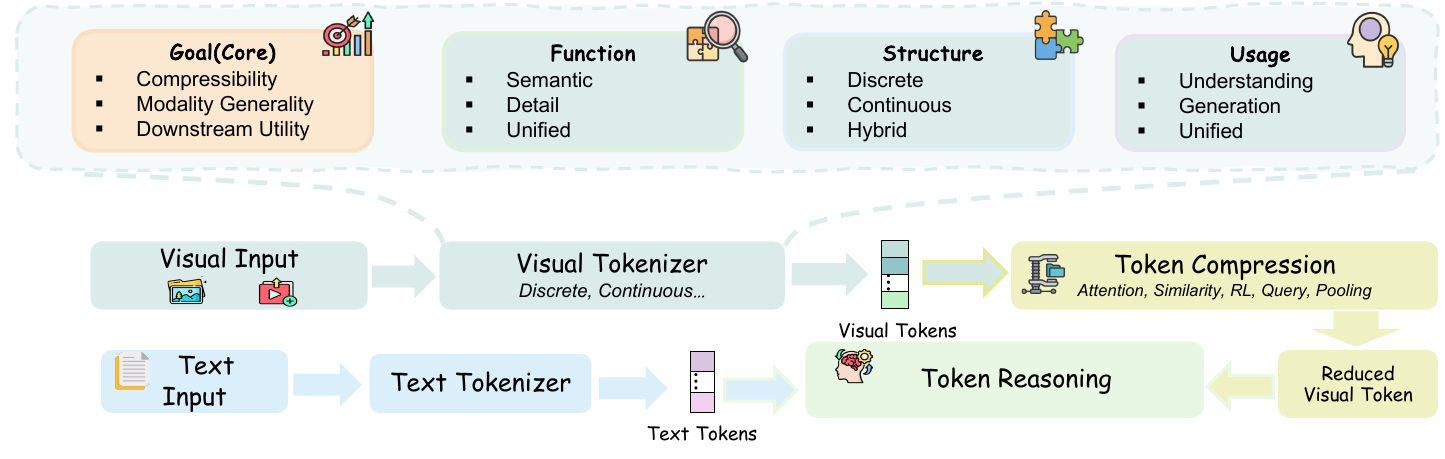}
    \vspace{-20pt}
    \caption{\textbf{Pipeline of (visual) token technology, typically used in the mainstream (multi-modal) large language models (LLMs/MLLMs).} Visual inputs are first converted into \emph{visual tokens} by a visual tokenizer, which may be either \emph{continuous} (patchify + linear projection with positional encoding, as in CLIP/SigLIP/DINOv2) or \emph{discrete} (latent encoding and codebook quantization, as in VQ-VAE/VQ-GAN), thereby forming transformer-ready sequences. A subsequent \emph{visual token compression} stage (e.g., attention-, similarity-, query-, pooling-, or RL-based) reduces the visual tokens to a small budget that, together with text tokens, feeds the \emph{token reasoning} module for cross-modal fusion and inference. Arrows indicate the data flow from tokenization to compression, then reasoning.}

\vspace{-15pt}

    \label{fig:token overview}
\end{figure*}
\subsection{Overview}

We organize \emph{visual token technologies} into three stages of the multimodal pipeline (Fig.~\ref{fig:token overview}): 1).  \textbf{visual tokenization}; 2).\textbf{visual token compression}; and 3).\textbf{cross-modal fusion and reasoning}. Tokenization is either \emph{continuous}, mapping images/videos to patch- or region-level embeddings for attention backbones, or \emph{discrete}, quantizing latents into codebook ``words'' that form compact symbolic sequences amenable to generative and autoregressive modeling. Because visual tokens usually dominate sequence length, compression reduces $N$ to $K$ ($K\!\ll\!N$), improving latency/throughput, reducing memory and KV-cache, and expanding spatial--temporal context under fixed compute. Importantly, $K$ is an explicit \emph{interface constraint}: compressors expose at most $K$ tokens, and downstream connectors, query bottlenecks, attention patterns, and decoders are designed and evaluated under this fixed allowance. In Sec.~\ref{sec:compression}, we systematize compressors by \emph{objective}, \emph{mechanism}, \emph{training regime}, \emph{location}, \emph{guidance}, and \emph{schedule}, and detail complexity and memory trade-offs. With fixed $K$, cross-modal reasoning scales as $\mathcal{O}(MK)$ per layer for $M$ text tokens. We highlight three operating modes: \emph{understanding} (image/video $\rightarrow$ text) via lightweight connectors or learned queries, \emph{generation} (text $\rightarrow$ image/video or editing) via AR or hybrid AR--diffusion decoders conditioned on $K$ tokens, and \emph{unified} models that read and emit visual tokens in an interleaved sequence. This framing links upstream tokenization to downstream reasoning through the compression budget, motivating the design choices and evaluation criteria developed next.

\begin{table*}[t] 
\centering 
\caption{Representative \emph{visual token compression} methods positioned along six axes (compact labels defined above).} 
\label{tab:vtoken-15x7-categorical} \small \setlength{\tabcolsep}{3.2pt} 
\begin{tabular}{@{}p{0.18\linewidth} p{0.11\linewidth} p{0.13\linewidth} p{0.095\linewidth} p{0.11\linewidth} p{0.11\linewidth} p{0.10\linewidth}@{}} \toprule \textbf{Method (abbr.)} & \textbf{Goal} & \textbf{Mechanism} & \textbf{Training} & \textbf{Location} & \textbf{Guidance} & \textbf{Schedule} \\ 
\midrule 
DeepSeek-OCR~\cite{wei2025deepseek} & Long. & Trans. & E2E & Bridge & Vision & Static \\ 
\midrule 
VoCo-LLaMA~\cite{ye2025voco} & Accel+Mem/KV & Query & Post. & LM & Vision & Static \\
\midrule 
DynamicViT~\cite{rao2021dynamicvit} & Accel. & Attn. & E2E & Enc. & Vision & Dynamic \\
\midrule 
FastV~\cite{chen2024image} & Accel+Mem/KV & Attn. & TF & LM & Text & One-shot \\ 
\midrule 
IVTP~\cite{huang2024ivtp} & Accel. & Attn. & Post. & LM & Text & Dynamic \\ 
\midrule 
VisionZip~\cite{yang2025visionzip} & Accel. & Attn. & Post. & Bridge & Vision & One-shot \\ 
\midrule 
PruneVid~\cite{huang2025prunevid} & Accel+Long & Sim. & TF & Hybrid & Text & Dynamic \\ 
\midrule 
HoliTom~\cite{shao2025holitom} & Accel. & Sim+Attn & TF & Hybrid. & Vision & Dynamic \\ 
\midrule 
RL4EViT~\cite{lu2025reinforcement} & Accel. & RL & RL & Enc. & Vision & Dynamic \\ 
\midrule 
VisionThink~\cite{yang2025visionthink} & Accel. & RL & RL & Enc. & Text & Dynamic \\ 
\midrule 
VisPruner~\cite{zhang2025vispruner} & Accel. & Attn+Sim & TF & LM & Vision & One-shot \\ 
\midrule 
DivPrune~\cite{alvar2025divprune} & Accel. & Sim & TF & LM & Vision & One-shot \\ 
\midrule 
LLaVA-UHD~\cite{guo2024llava} & Accel+Long & Trans+Query & E2E & Bridge & Vision & Static \\ 
\bottomrule \end{tabular} 
\vspace{3pt} 
\\
\footnotesize \textit{Notes:} (i) When methods plausibly serve multiple objectives, we list the \emph{primary} ones. (ii) “One-shot” denotes a single pruning/compression step during prefill; “Prog.” denotes layer-wise progressive merging/pruning. (iii) “Hybrid” under \emph{Location} indicates coordinated outer-LLM (pre-/post-encoder) and inner-LLM stages. 
\vspace{-15pt}
\end{table*}

\subsubsection{Visual Tokenization}

\noindent \textit{Visual tokenization} converts images/videos into transformer-compatible token sequences~\cite{vaswani2017attention}. Methods largely fall into two families. \emph{Continuous} tokenizers partition inputs into patches (or regions) and map each unit to an embedding via linear projection with positional encoding, as in ViT-style backbones and representation learners such as CLIP~\cite{radford2021clip}, SigLIP~\cite{tschannen2025siglip}, and DINOv2~\cite{oquab2024dinov2}. \emph{Discrete} (codebook-based) tokenizers encode inputs into latents and quantize them into codebook indices, as in VQ-VAE~\cite{oord2017vqvae} and VQ-GAN~\cite{esser2021taming}; the resulting sequences are typically modeled with autoregressive or diffusion priors. Continuous tokenizers dominate perception/understanding pipelines and supply the main compressible visual inputs to LVLMs~\cite{wang2024qwen2}, whereas discrete tokenizers provide learned visual vocabularies central to high-fidelity generation. From an information-theoretic view, both can be cast as learned source coding: a transform stage (e.g., convolutional~\cite{o2015introduction} or linear~\cite{tolstikhin2021mlp} encoders), a quantization/selection stage (e.g., codebooks~\cite{zhang2024conceptual} or structured resampling~\cite{alayrac2022flamingo}), and a probabilistic modeling stage (e.g., LLMs~\cite{minaee2024large} or diffusion models~\cite{yang2023diffusion}). This lens links token formation to downstream fusion and generation, supporting more unified evaluation across understanding- and generation-centric systems.

\subsubsection{Visual Token Compression}
\label{sec:compression}
\begin{figure*}[t]
    \centering
    \vspace{-2mm}
    \includegraphics[width=0.8\linewidth]{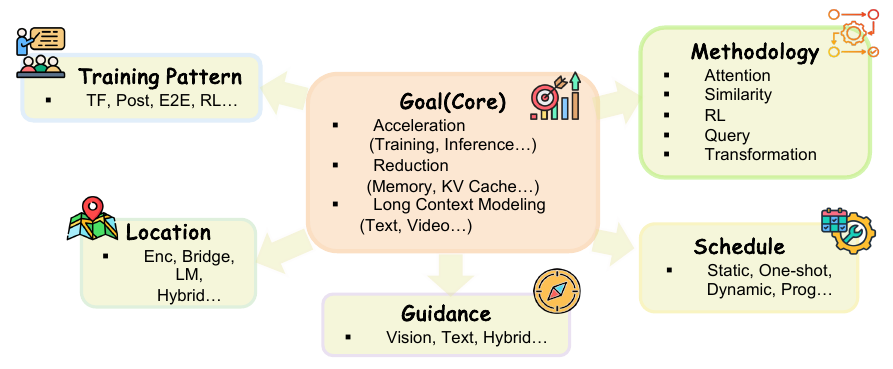}
    \vspace{-10pt}
    \caption{\textbf{Six–axis view of visual token compression.} The center \emph{Goal} (acceleration, memory/KV reduction, long-context) is realized by choices along five orthogonal axes: \emph{Methodology} (attention, similarity, RL, query, transformation), \emph{Training pattern} (TF, Post, E2E, RL), \emph{Location} (encoder, bridge, LM/KV, hybrid), \emph{Guidance} (vision, text, hybrid), and \emph{Schedule} (static, one-shot, dynamic, progressive). Arrows indicate how these factors compose into a concrete compression policy under a fixed visual budget $K$; the taxonomy matches Sec.~\ref{sec:compression} and Table~\ref{tab:vtoken-15x7-categorical}.}
    \vspace{-10pt}
    \label{fig:token compression}
\vspace{-8pt}
\end{figure*}

\textbf{Problem setup \& scope.}
In LVLMs~\cite{guo2024llava, liu2024llavanext, wang2025internvl3}, visual tokens often dominate the multimodal sequence. For an $H{\times}W$ image with patch size $p{\times}p$, $N_{\text{img}}=\frac{HW}{p^{2}}$; for a $T$-frame video, $N_{\text{vid}}=T\cdot\frac{HW}{p^{2}}$. This growth rapidly exhausts attention and KV-cache budgets, becoming a primary latency/memory bottleneck. \emph{Visual token compression} maps an original set of $N$ visual tokens to a smaller, task-faithful set of $K$ ($K\!\ll\!N$), reducing per-layer cross-attention from $\mathcal{O}(MN)$ to $\mathcal{O}(MK)$ (for $M$ active text tokens) and vision self-attention from $\mathcal{O}(N^{2})$ to $\mathcal{O}(K^{2})$, while retaining the information needed for fusion and reasoning.

\textbf{Goal (why compress).}
We emphasize three objectives. \emph{\textbf{Accel.}} reduces wall-clock time and FLOPs by shrinking the active visual sequence (e.g., merging/pruning), improving prefill cost (e.g., ToMe~\cite{bolya2022tome}). \emph{\textbf{Mem/KV}} targets peak VRAM and KV-cache footprint via early, decode-consistent reduction (e.g., TopV~\cite{yang2025topv}), remaining compatible with FlashAttention~\cite{dao2023flashattention2}/KV paging. \emph{\textbf{Long.}} extends admissible spatial--temporal context under fixed resources, e.g., fixed per-frame budgets in long-video LVLMs (LLaMA-VID~\cite{li2024llama}).

\textbf{Mechanism (how to compress).}
We group techniques into five families that trade a large visual token set for a compact representation. \emph{\textbf{a. Similarity-based.}} Merge redundant tokens to preserve coverage with fewer representatives: canonical ToMe-style merging~\cite{bolya2022tome}, late-stage merging (FOLDER~\cite{wang2025folder}), diversity-driven subset selection (DivPrune~\cite{alvar2025divprune}), and stage-wise fusion (AuroraCap~\cite{chai2024auroracap}); video variants combine spatial merging with temporal redundancy control (Chat-UniVi~\cite{jin2024chat}, FastVID~\cite{shen2025fastvid}, DynTok~\cite{zhang2025dyntok}). \emph{\textbf{b. Attention-based.}} Use attention-derived salience to keep informative tokens and prune the rest, including encoder-side hierarchical schemes (VisPruner~\cite{zhang2025vispruner}, MustDrop~\cite{liu2024multi}, VScan~\cite{zhang2025vscan}, HiReD~\cite{arif2025hired}, GlobalCom$^{2}$~\cite{liu2025global}) and LVLM-side, layer-aware schedules or learned thresholds (PyramidDrop~\cite{xing2024pyramiddrop}, VTW~\cite{lin2025boosting}, FitPrune~\cite{ye2025fit}, ST$^{3}$~\cite{zhuang2025st3}, ATP-LLaVA~\cite{ye2025atp}, ZipVL~\cite{he2024zipvl}), often preserving VQA/Caption quality with small $K$. \emph{\textbf{c. Query-based.}} Replace dense tokens with a fixed-$K$ bottleneck that reads vision features via cross-attention and forwards only summaries to the LM, as in Flamingo~\cite{alayrac2022flamingo} and BLIP-2~\cite{li2023blip2}, with extensions such as InstructBLIP~\cite{dai2023instructblip}, mPLUG-Owl~\cite{ye2023mplug}, MiniGPT-4~\cite{zhu2024minigpt}, Victor~\cite{wen2024efficient}, and extreme pre-fusion interfaces (LLaVA-Mini~\cite{zhang2025llava}); video instantiations constrain tokens per frame or distill into LM-internal codes (BLIP-3-Video~\cite{ryoo2024xgen}, Long-VMNet~\cite{gurukar2025long}). With architectural $K$, fusion cost is near-constant as resolution or $T$ grows. \emph{\textbf{d. Transformation-based.}} Reduce tokens at ingress by changing the sampling before fusion (downsampling, pyramids, learned pooling), e.g., pooling-based front-ends (LLaVA-OneVision~\cite{li2024llava}, DeCo~\cite{yao2024deco}), multi-granular pooling (M$^{3}$~\cite{cai2024matryoshka}), lightweight conv abstractors (Honeybee C-Abstractor~\cite{cha2024honeybee}, MobileVLM LDP~\cite{chu2023mobilevlm}), and tiling with compression heads for UHD inputs (NVLM~\cite{dai2024nvlm}). \emph{\textbf{e. RL-based.}} Formulate compression as budgeted decision-making with rewards balancing fidelity and efficiency: multi-agent layer-wise pruning (RL4EViT~\cite{lu2025reinforcement}), event-triggered streaming controllers (MARC~\cite{wu2025marc}), and instance-adaptive policies for difficult inputs (VisionThink~\cite{yang2025visionthink}). 

\textbf{Training (whether to train).} \emph{\textbf{TF}} methods are plug-and-play (PruMerge~\cite{shang2024prumerge}, SparseVLM~\cite{zhang2024sparsevlm}, LLaVA-Scissor~\cite{sun2025llava}). \emph{\textbf{Post}} tunes only lightweight selectors/bridges while freezing backbones (TokenPacker~\cite{li2024tokenpacker}, VideoChat-Flash~\cite{ma2025videochatflash}). \emph{\textbf{E2E}} co-optimizes intake/projectors so the model natively operates at small $K$ (LLaVA-NeXT~\cite{liu2024llavanext}, VideoLLaMA-2~\cite{zhang2024videollama2}, TimeChat~\cite{ren2024timechat}). \emph{\textbf{RL}} learns budget-aware policies with explicit efficiency--accuracy rewards (MARC~\cite{wu2025marc}, RL4EViT~\cite{lu2025reinforcement}).

\textbf{Location $\mathcal{L}$ (where to act).}
\emph{\textbf{Enc.}} compresses before fusion (PatchMerger~\cite{renggli2022patchmerger}); \emph{\textbf{Bridge}} retokenizes to a compact interface (Kosmos2-style connectors~\cite{peng2023kosmos2}, Perceiver-IO resampling~\cite{jaegle2021perceiverio}); \emph{\textbf{LM}} prunes inside the decoder (SparseVLM~\cite{zhang2024sparsevlm}); \emph{\textbf{KV}} directly controls cache growth (VL-Cache~\cite{tu2024vlcache}, LOOK-M~\cite{wan2024lookm}); \emph{\textbf{Hybrid}} coordinates stages for large reductions (VideoChat-style systems~\cite{maaz2024video}).

\textbf{Guidance (who guides).}
\emph{\textbf{Vision guidance}} uses bottom-up cues (vid-TLDR~\cite{choi2024vidtldr}, VisionDrop~\cite{zhang2025visiondrop}); \emph{\textbf{Text guidance}} conditions selection on instructions (LVPruning~\cite{sun2025lvpruning}, CTFP~\cite{luo2025large}); \emph{\textbf{Hybrid}} combines both (PTP~\cite{liang2025ptp}, TCR~\cite{korbar2024tcr}).

\textbf{Schedule (when compress).}
\emph{\textbf{Static}} uses fixed ratios/budgets (TCR~\cite{korbar2024tcr}, vid-TLDR~\cite{choi2024vidtldr});
\emph{\textbf{One-shot}} prunes once at prefill for decode-consistent savings (LLaVA-Scissor~\cite{sun2025llava});
\emph{\textbf{Dynamic}} adapts budgets per input/prompt (LVPruning~\cite{sun2025lvpruning}, Dynamic-VLM~\cite{liu2025dynamicvlm});
\emph{\textbf{Progressive}} sparsifies across layers/depths (CoViPAL~\cite{xu2025covipal}, FEATHER~\cite{endo2025feather}).

\subsubsection{Cross-Modal Token Fusion and Reasoning}
\textbf{Problem framing.}
With a compressed visual budget of $K$ tokens, fusion must present these tokens to the LLM so cross-modal interactions scale as $\mathcal{O}(MK)$ per layer, where $M$ is the active text length. Architectures mainly differ in (i) how visual tokens enter the decoder (connectors, learned query bottlenecks, or two-stream encoders) and (ii) how the decoder reasons over mixed evidence (attention, grounded pointers, or program/tool execution).

\textbf{Understanding (image/video $\rightarrow$ text).}
Two-stream encoders (ViLBERT~\cite{lu2019vilbert}, LXMERT~\cite{tan2019lxmert}) established cross-attention blueprints for vision--language alignment. Modern LVLMs either (a) project vision features into the LLM token space with lightweight connectors (e.g., LLaVA~\cite{liu2023visual,liu2023improvedllava}) or (b) aggregate dense features via learned query bottlenecks before the LLM (Flamingo~\cite{alayrac2022flamingo}, BLIP-2~\cite{li2023blip2}). For video, pre-projection alignment can stabilize fusion under small per-frame budgets (Video-LLaVA~\cite{lin2024video}). Grounded pointers such as location tokens (Kosmos-2~\cite{peng2023kosmos2}) make spatial references explicit, while program/tool routes (ViperGPT~\cite{suris2023vipergpt}) externalize long reasoning when attention alone is insufficient.

\textbf{Generation (text $\rightarrow$ image/video, or editing).}
Autoregressive (AR) generators interleave text and visual tokens in a single next-token stream (CM3LeOn~\cite{yu2023cm3leon}), supporting prompting, infilling, and editing. Hybrid AR+diffusion designs retain one Transformer trunk but attach diffusion-style heads for higher-fidelity synthesis (Show-o~\cite{xie2024showo}), allowing a shared backbone across understanding and generation. Strong visual priors can further improve conditional generation when the model is constrained to an informative budget of $K$ tokens (Emu~\cite{sun2023emu}).

\textbf{Unified understanding \& generation (U\&G in one model).}
Unified models adopt a shared token space/objective so one network both \emph{reads} and \emph{emits} visual tokens. Purely autoregressive formulations perform next-token prediction over interleaved text and quantized visual tokens (VILA-U~\cite{wu2024vilau}, Chameleon~\cite{team2024chameleon}). Scaling decoder-only pretraining on interleaved corpora further improves both capabilities and induces stronger multimodal reasoning (BAGEL~\cite{deng2025emerging}). Hybrid designs combine autoregressive decoding with flow matching or diffusion-style generation under a shared Transformer trunk (Show-o2~\cite{xie2025show}). Inference attends jointly to $K$ visual tokens and $M$ text tokens, enabling grounded instruction following and multi-step reasoning without separate understanding/generation models.

\subsection{Architectures of visual tokenizers}
\noindent Visual tokenizers determine the mechanism by which raw visual signals are converted into latent representations that are suitable for downstream processing, including both generation and understanding tasks. Architecturally, they convert a spatially structured signal into a sequence of tokens that align with the transformer or diffusion interface. Current designs fall into two categories, \emph{continuous} and \emph{discrete} tokenizers, each characterized by distinct design philosophie, objectives, and downstream compatibilities.

\subsubsection{Continuous tokenizers}
Continuous tokenizers, also known as vision encoder, are designed to convert visual patches into continuous embedding vectors via vision transformers and linear projections, incorporating positional encodings. These embeddings remain differentiable and support gradient-based optimization throughout multimodal pretraining. During the visual understanding era, they have become the predominant choice in MLLMs for image and video understanding, including representative approaches like the vision encoders of LLaVA\cite{liu2023visual}. The tokenizer produces high-dimensional latent sequences whose embeddings are compatible with those of text embeddings for direct ingestion by the LLM.

Continuous visual tokenizers form the architecture of nearly all modern MLLMs. Their design inherits from ViT\cite{ViT} and contrastive representation learners such as CLIP\cite{radford2021clip} and SigLIP\cite{tschannen2025siglip}. A typical tokenizer consists of three stages: (i) \textbf{patchification}, splitting the image into non-overlapping patches of size $P\times P$; (ii) \textbf{projection}, flattening and linearly mapping each patch into a $d$ dimensional embedding; and (iii) \textbf{positional encoding}, injecting spatial or temporal information. The resulting sequence of embeddings $X_V \in \mathbb{R}^{N\times d}$, where $N = H\times W/P^2$.

In MLLMs, the tokenizer is typically initialized from a pre-trained visual encoder (e.g., CLIP ViT-L/14) then frozen or fine-tuned. Continuous tokens are aligned with text embeddings thus semantically rich. Representative approaches include LLaVA\cite{liu2023visual}, which relies on a frozen CLIP encoder and a subsequent projection module for cross-modal alignment; BLIP-2\cite{li2023blip2}, which builds its vision tower upon a frozen CLIP encoder with Q-Former; and Qwen2-VL\cite{wang2024qwen2}, which introduces a dynamic-resolution vision transformer for flexible visual encoding. In the video domain, methods such as VideoLLaMA\cite{zhang2023video} and InternVideo\cite{wang2022internvideo} extend image-based encoders with spatio-temporal patching and using a projection modules to maintain alignment across frames.  
Continuous tokenizers are differentiable and easily integrated into multimodal backbones, but they may produce redundancy at high resolution and lack explicit interpretability.

\subsubsection{Discrete tokenizers}
Discrete tokenizers, in contrast, encode visual inputs into indices of a learned codebook, thereby producing a symbolic token sequence. This quantization process such as in VQ-VAE\cite{oord2017vqvae}, VQ-GAN\cite{esser2021taming}, bridges perception and generation: each index corresponds to a visual word learned from the data. In diffusion or autoregressive generative models, such discrete tokenizers serve as visual vocabulary or compressesor. In MLLMs, discrete tokenization is less commonly used for understanding tasks, but is typically adopted in unified modeling frameworks that represent both images and text as discrete sequences under a shared embedding space.

Discrete visual tokenizers initially designed for generative models, have recently gained attention in multimodal understanding and unified modeling. Their core idea is to represent an image $x$ by a sequence of quantized codes $\{z_i\}_{i=1}^N$, where each $z_i \in \{1,\dots,K\}$ indexes a learned codebook $\mathcal{C} \in \mathbb{R}^{K\times d}$. The encoder $E_\phi$ maps $x$ to latent features, which are then quantized to the nearest codebook entry:  
\[
z_i = \arg\min_{k}\|E_\phi(x)_i - \mathcal{C}_k\|_2^2, \quad
\hat{x} = D_\psi(\mathcal{C}_{z_1}, \ldots, \mathcal{C}_{z_N}).
\]
This pipeline, introduced by VQ-VAE and refined by VQ-GAN, forms the foundation of many discrete vision language systems.

In MLLMs, discrete tokenizers treats images as symbolic sequences as the text do. For instance, CM3\cite{yu2023cm3leon}, MAGVIT2\cite{Magvit-v2} unify text and images modalities by mapping images into discrete tokens and jointly training a transformer to autoregress over text–image sequences. This enables bidirectional tasks such as captioning, visual question answering, and text-to-image generation within a single language backbone. Compared to continuous tokenizers, discrete tokenizers demonstrate superior compatibility with LLM architecture, produce more compact representation, but require non-differentiable quantization steps and codebook maintenance. They also tend to lose fine-grained details critical for dense reasoning.
\emph{Overall}, continuous tokenizer is more lossless and suitable for understanding tasks, dominate current MLLM architectures for reasoning. Discrete tokenizers is more efficient and compatible for unified generative frameworks. The boundary between them is increasingly blurred as hybrid systems adopt quantized continuous latents or learnable patch embeddings jointly optimized with language supervision. 


\begin{table*}[t] 
\centering
\small


\caption{Representative visual tokenizers, classified by the model architectures and target tasks they were designed for.}
\begin{tabular}{lcccc}
\toprule
\textbf{Method} & \textbf{Continuous / Discrete} & \textbf{MLLM / Diffusion} & \textbf{Generation / Understanding} \\
\midrule
CLIP\cite{radford2021clip} & Continuous & MLLM & Understanding \\
SigLIP\cite{tschannen2025siglip} & Continuous & MLLM & Understanding \\
BLIP-2\cite{li2023blip2} & Continuous & MLLM & Understanding \\
LLaVA\cite{liu2023visual} & Continuous & MLLM & Understanding \\
Qwen2-VL\cite{bai2025qwen2} & Continuous & MLLM & Understanding \\
VideoLLaMA\cite{zhang2023video} & Continuous & MLLM & Understanding \\
InternVideo\cite{wang2022internvideo} & Continuous & MLLM & Understanding \\ 
\midrule
VQ-VAE\cite{oord2017vqvae} & Discrete & Diffusion / MLLM & Generation \\
VQ-GAN\cite{esser2021taming} & Discrete & Diffusion / MLLM & Generation \\
DALL$\cdot$E\cite{ramesh2021dalle} & Discrete & MLLM & Generation \\
MaskGIT\cite{chang2022maskgit} & Discrete & MLLM & Generation \\
MAGVIT2\cite{Magvit-v2} & Discrete & MLLM & Both Gen \& Understanding \\
CM3\cite{yu2023cm3leon} & Discrete & MLLM & Both Gen \& Understanding \\
\midrule
LDM\cite{rombach2022ldm} & Continuous & Diffusion & Generation \\
REPA\cite{yu2024representation} & Continuous & Diffusion & Generation \\
RAE\cite{zheng2025diffusion} & Continuous & Diffusion & Generation \\
\bottomrule
\end{tabular}
\vspace{-10pt}
\end{table*}

\subsection{Generation Task}

\begin{figure}[!t]
\centering
\vspace{-20pt}
\includegraphics[width=0.7\linewidth, angle=270]{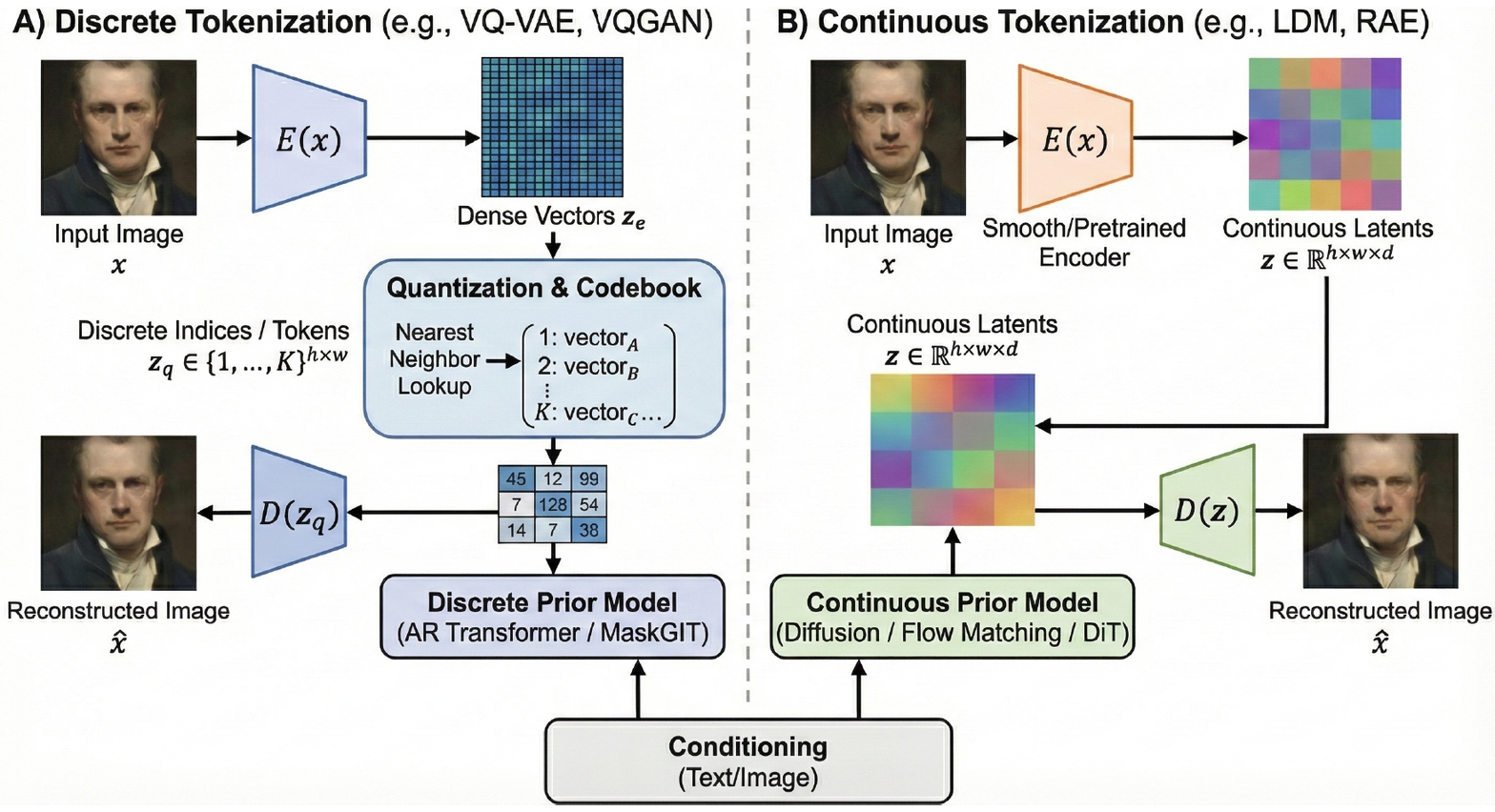}
\vspace{-20pt}
\caption{Comparison of Discrete vs. Continuous Image Tokenization Paradigms. \textbf{(A) Discrete Tokenization:} The input image is encoded into dense vectors and then quantized using a learnable codebook to produce discrete indices ($z_q$). These tokens are processed by a discrete prior model (e.g., AR \cite{yu2024image} Transformer \cite{vaswani2017attention}). \textbf{(B) Continuous Tokenization:} The encoder maps the image directly to continuous latent variables ($z$) without quantization. These latents are modeled by a continuous prior, such as a Diffusion Model \cite{ho2020denoising} or Flow Matching \cite{lipman2022flow}. Both frameworks utilize conditioning inputs (e.g., text) and a decoder for image reconstruction.}
\label{fig:gen}
\vspace{-18pt}
\end{figure}

To situate the discussion, Fig.~\ref{fig:gen} compares the two dominant paradigms. Discrete tokenization (Fig.~\ref{fig:gen}A) maps encoder features to finite codebook indices, suitable for autoregressive or masked priors. Continuous tokenization (Fig.~\ref{fig:gen}B) maps images to smooth latent spaces, enabling diffusion or flow-based priors. This dichotomy underpins modern generative pipelines.

Discrete tokenization quantizes images into index grids. VQ-VAE enabled visual sequence modeling \cite{oord2017vqvae}, while VQGAN enhanced fidelity via adversarial objectives \cite{esser2021taming}. These naturally pair with discrete priors: DALL·E employs autoregressive transformers for open-vocabulary generation \cite{ramesh2021dalle}, whereas MaskGIT uses masked parallel decoding to improve efficiency \cite{chang2022maskgit}. Recent advances like TiTok and VAR further optimize throughput for high resolutions \cite{yu2024image,tian2024visual}. While discrete formulations offer entropy coding, they suffer from codebook pathologies and sequence inflation at high resolutions, requiring hierarchical mitigations.

Continuous tokenization replaces quantization with smooth, expressive latents. RAEs simplify training by pairing frozen pretrained encoders with lightweight decoders \cite{zheng2025diffusion}, and REPA improves DiT-style models via feature alignment \cite{yu2024representation}. Stable Diffusion performs denoising in perceptually regularized latent spaces \cite{rombach2022ldm}, supported by accelerated samplers. The differentiable nature of continuous latents facilitates gradient-based editing and parameter-efficient adaptation. However, the absence of quantization complicates exact bit-level accounting, and sampling costs can dominate without accelerated objectives.

Across both families, controllability is achieved via cross-modal connectors or structural branches, which enforce geometry and layout \cite{zhang2023controlnet}, while video extensions utilize spatiotemporal grids and consistency losses to maintain coherence. The choice represents a trade-off: discrete tokenization offers compactness and probabilistic clarity at the cost of quantization artifacts, whereas continuous tokenization prioritizes smoothness and editability at the expense of sampling complexity. Hybrid designs are increasingly adopted to combine the structural advantages of discrete compressibility with the alignment benefits of continuous spaces.

\subsection{Understanding Task}

\begin{figure}[th]
\centering
\vspace{-10pt}
\includegraphics[width=1\linewidth]{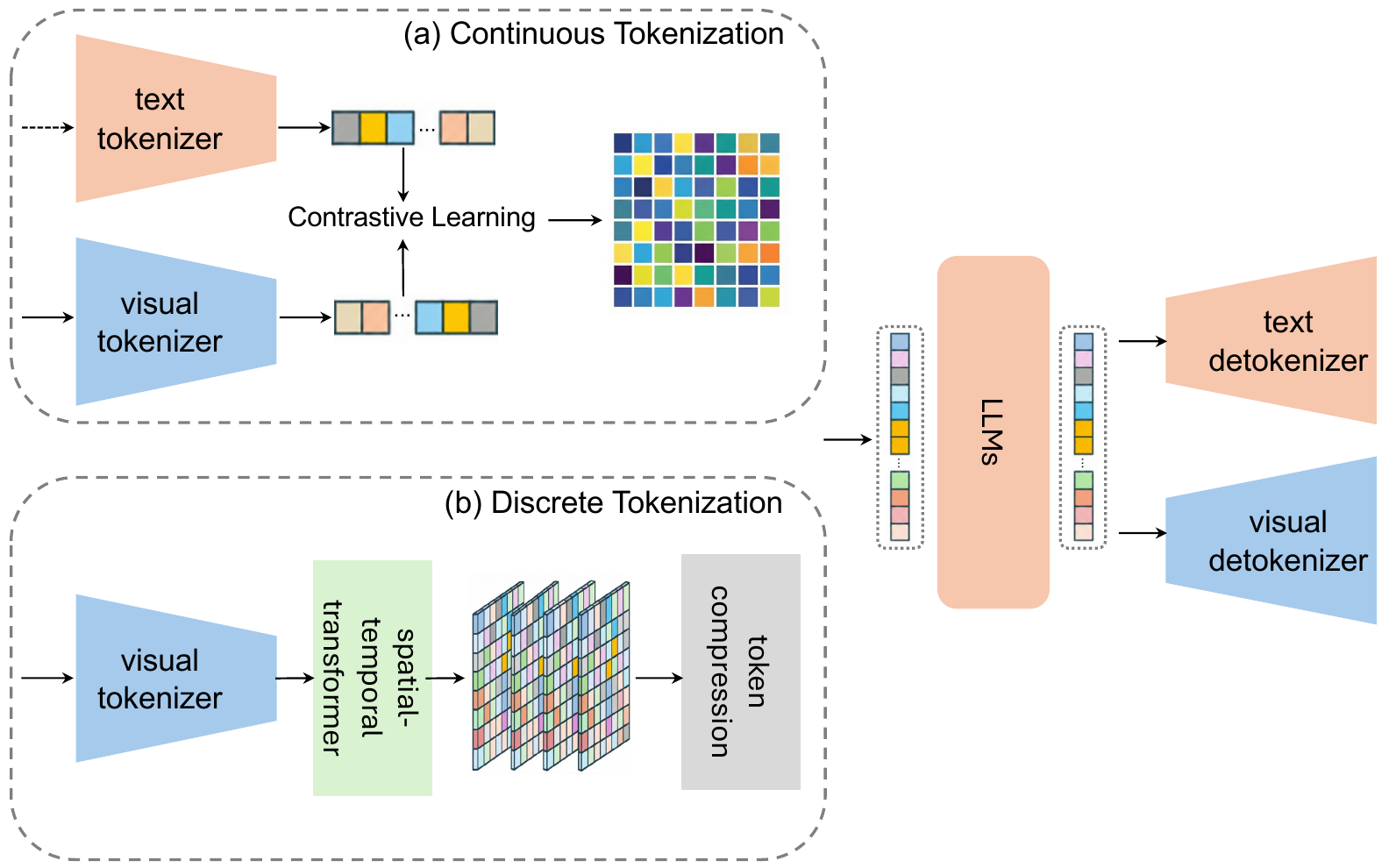}
\vspace{-20pt}
\caption{Visual token technologies for understanding tasks. (a) Continuous visual tokenization~\cite{ViT,radford2021clip,tschannen2025siglip}represents visual inputs using compact latent embeddings. (b) Discrete visual tokenization~\cite{oord2017vqvae,yu2023magvit,Magvit-v2,ge2022long_TATS} encodes images and videos into compact symbolic tokens. Integrated token compression~\cite{zhao2024_CV-VAE,wang2024_OmniTokenizer,gao2024linvt} complements both by controlling token quantity for scalable multimodal reasoning.}
\label{fig:understanding}
\vspace{-18pt}
\end{figure}

\subsubsection{Motivation}
Visual images and videos contain rich spatial and temporal information but also substantial redundancy, making direct pixel-level modeling inefficient. To address this, recent work focuses on compact visual token representations that transform raw pixels into low-dimensional latent tokens, preserving essential semantics and dynamics while discarding irrelevant details. As shown in figure~\ref{fig:understanding}, an encoder typically converts inputs into spatial or spatio-temporal tokens for downstream understanding. Within this paradigm, two challenges arise: defining effective representational units and controlling token quantity for scalable reasoning. Together, these enable efficient visual understanding based on compact, semantically meaningful tokens rather than dense pixels.

\subsubsection{Compact Tokenization and Compression for Visual Understanding}

The development of compact visual tokens forms a continuum from representation learning to efficient reasoning.
Early visual transformers such as ViT~\cite{ViT} and CLIP~\cite{radford2021clip} represented images as fixed-size patch embeddings, which captured spatial structure but produced hundreds of redundant tokens per frame. Subsequent research introduced learned tokenizers—such as VQ-VAE~\cite{oord2017vqvae}, VQGAN~\cite{esser2021taming}, and MAGVIT~\cite{yu2023magvit}—that replaced raw embeddings with discrete codebook indices, offering compact symbolic representations compatible with transformer and language-model-based architectures. These methods achieved strong perceptual compression while preserving object and texture semantics. Continuous and hybrid designs, such as CV-VAE~\cite{zhao2024_CV-VAE}, OmniTokenizer~\cite{wang2024_OmniTokenizer}, and BSQ-ViT~\cite{zhao2024BSQ-ViT}, further improved reconstruction fidelity and cross-modal alignment by encoding fine-grained visual context into continuous latent spaces. For visual understanding, these compact image tokens act as semantic building blocks, capturing scene layout and entity relationships in a format readily interpretable by multimodal reasoning models.

Extending spatial compression into the temporal dimension introduces new challenges—motion coherence, temporal redundancy, and causal consistency. Early video extensions of image tokenizers simply applied 2D encoders frame by frame, resulting in redundant and temporally inconsistent tokens. To overcome this, models such as TATS~\cite{ge2022long_TATS}, MAGVIT-V2~\cite{Magvit-v2}, and CogVideoX~\cite{yang2024cogvideox} employ spatio-temporal quantization, learning shared codebooks or latent VAEs that jointly model spatial appearance and temporal dynamics. Diffusion-oriented systems like OpenSora~\cite{zheng2024opensora}, CV-VAE~\cite{zhao2024_CV-VAE}, and HunyuanVideo~\cite{kong2025hunyuanvideo} further compress videos into continuous latent sequences suitable for generative or reasoning backbones. These tokenizers aim for semantic sufficiency—capturing key entities, motions, and interactions—rather than pixel-perfect detail, which is redundant for understanding tasks. Hybrid tokenizers (e.g., LinVT~\cite{gao2024linvt}, TVC~\cite{zhou2025tvc}) explicitly balance reconstruction quality with token compactness by combining discrete quantization with continuous temporal compression, providing representations that are both efficient and expressive across modalities.

Even with efficient tokenizers, the total number of tokens in long or high-resolution videos often exceeds the processing limits of large transformers. To mitigate this, compression mechanisms are increasingly integrated into tokenization pipelines. Transformation-based methods (e.g., PLLaVA~\cite{xu2024PLLaVA}, VideoLLaMA-2~\cite{zhang2024videollama2}) employ learnable pooling or convolution layers to summarize local regions or temporal segments into fewer tokens while preserving coarse semantics. Similarity-based clustering (e.g., Chat-UniVi~\cite{jin2024chat}, FastVID~\cite{shen2025fastvid}, HoliTom~\cite{shao2025holitom}) merges highly correlated frame or patch tokens based on feature proximity, reducing redundancy while maintaining contextual continuity. Attention-guided pruning (e.g., VisionZip~\cite{yang2025visionzip}, FastV~\cite{chen2024image}) removes low-saliency tokens using attention scores or importance maps, while query-driven selection (e.g., Token Turing Machines~\cite{ryoo2023tokenturingmachines}, BLIP-3-Video~\cite{ryoo2024xgen}, LongVU~\cite{shen2024longvu}) selectively retains tokens relevant to a textual or task-driven query. In practice, these techniques are often jointly optimized with the encoder, producing adaptive compression that responds to content and task complexity.

Across benchmarks~\cite{yu2019ActivityNet-QA,li2024mvbench,fu2025videomme}, studies show that retaining only 25–35\% of tokens preserves over 95\% of reasoning accuracy, indicating that most pixel-level details are redundant for semantic understanding. Current research therefore focuses on task-aware and adaptive tokenization, dynamically allocating representational capacity to semantically important regions or temporal segments.

\vspace{-0.3em}

\subsection{Unified Tokenizer}
\label{sec:unified_tokenizer}

\begin{figure}[t]
    \centering
    \includegraphics[width=\linewidth]{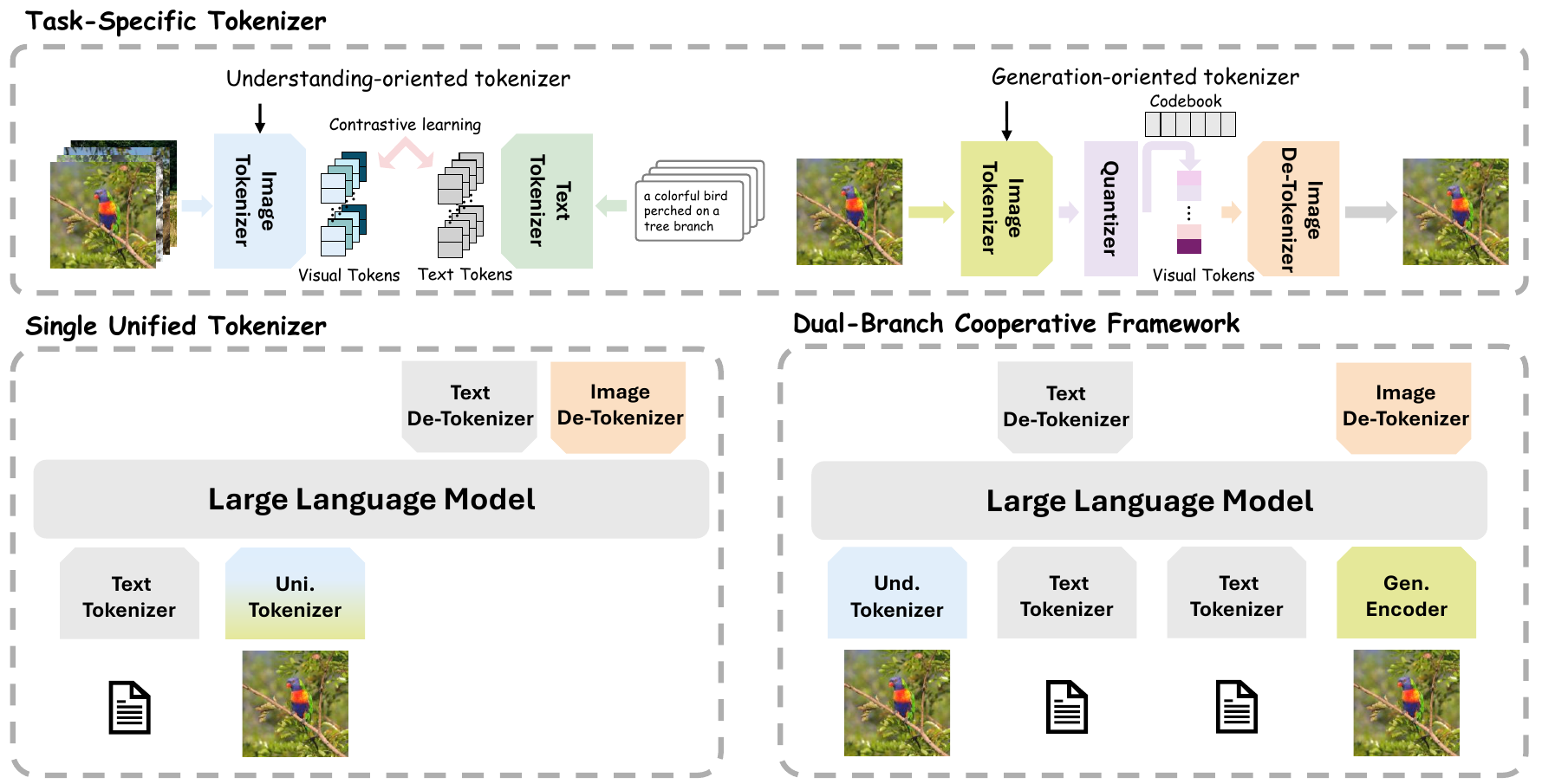}
    \caption{Overview of task-specific tokenizers (understanding-oriented~\cite{radford2021clip,zhai2023sigmoid,bolya2025perception} and generation-oriented~\cite{oord2017vqvae,esser2021taming}), dual-branch cooperative frameworks~\cite{team2024chameleon,xie2024showo}, and single unified tokenizers~\cite{wu2024vilau,ma2025unitok,lu2025atoken}. This taxonomy illustrates the evolution from separated representations toward unified visual tokenization.}
    \vspace{-15pt}
    \label{fig:tokenizer_taxonomy}
\end{figure}
To contextualize the evolution of visual tokenization, Fig.~\ref{fig:tokenizer_taxonomy} summarizes three major paradigms: 
(1) task-specific tokenizers designed separately for understanding or generation, 
(2) dual-branch cooperative frameworks that combine both types of representations, and 
(3) unified tokenizers that aim to bridge semantic alignment and pixel-level reconstruction within a shared latent space.
 Next, we discuss each category in detail.

\subsubsection{\small{Task-Specific Tokenizer: Understanding vs. Generation}}
As discussed above, most existing visual tokenizers are task-specific.
One family, represented by understanding-oriented tokenizers (usually in continuous form) such as CLIP \cite{radford2021clip}, SigLIP \cite{zhai2023sigmoid,tschannen2025siglip}, Perception Encoder~\cite{bolya2025perception}, and DINO\cite{caron2021emerging,oquab2024dinov2}, is trained through image–text alignment and excels in multimodal reasoning tasks such as VQA\cite{antol2015vqa} and image captioning\cite{stefanini2022show}. However, the lack of Generation supervision leads to clear bottlenecks in image generation and editing\cite{chen2025unireal}.
Another family, represented by generation-oriented tokenizers (usually in discrete form) such as VQ-VAE\cite{oord2017vqvae} and VQ-GAN\cite{esser2021taming}, enables high-fidelity image synthesis, generation, and editing. Nevertheless, their latent spaces are not semantically aligned with language, which limits their generalization to cross-modal understanding tasks. In addition, these models often require large-scale joint training data to align the latent space with downstream models.

\subsubsection{Dual-Branch Cooperative Framework}
As multimodal large models continue to evolve toward unified understanding and generation, this dichotomy has become increasingly restrictive.
Recent works demonstrate two opposing tendencies.
Models such as chameleon\cite{team2024chameleon} and Show-o\cite{xie2024showo} adopt generation-oriented tokenizers that achieve strong reconstruction fidelity and detailed generation but lack semantic alignment and controllability in multimodal reasoning.
In contrast, Models such as BLIP3-o\cite{chen2025blip3} and EMU2\cite{sun2024generative} adopt understanding-oriented tokenizers. Some require the large language model to directly predict continuous visual embeddings that interact with a diffusion module, while others quantize semantic features through a codebook mechanism to support generation. Despite these efforts, such methods still suffer from representation distortion, modality mismatch, and semantic drift.

Beyond these two extremes, an emerging line of research explores \textbf{dual-branch cooperative frameworks} that jointly leverage both understanding- and generation-oriented tokenizers within a unified model.
Representative works such as Janus~\cite{wu2025janus} and BAGEL~\cite{deng2025emerging} adopt a hybrid design, where a understanding-oriented tokenizer and a generation-oriented tokenizer are employed in parallel.
While this design generally delivers stable performance, but several limitations remain.
First, maintaining two different types of tokenizers simultaneously greatly increases the number of visual tokens in the input sequence, leading to higher inference latency and memory overhead.
Moreover, since the two branches are often pretrained with different objectives, their latent distributions may gradually diverge, resulting in semantic–visual inconsistency.
Consequently, recent studies have begun to explore a new unified paradigm: employing a single \textbf{Unified Tokenizer} that achieves both semantic alignment and detail-preserving reconstruction within a shared latent space.

\subsubsection{Single Unified Tokenizers}
As one of the earliest attempts toward a unified visual tokenizer, VILA-U\cite{wu2024vilau} builds upon the VQ-VAE framework and introduces contrastive learning between discrete visual tokens and text tokens, thereby enabling both visual understanding and generation within a single model.
UniTok\cite{ma2025unitok} points out that this joint training paradigm is difficult to stabilize, as the two objectives often interfere with each other, leading to minimal improvement in understanding but substantial degradation in generation. Further analysis reveals that the issue does not stem from conflicting tasks, but rather from the limited expressiveness of the discrete token space, which fails to capture the semantics required for understanding. To address this, UniTok proposes a Multi-Codebook Quantization Mechanism that expands the codebook capacity and dimensionality to enhance the representational power of discrete features.
TokenFlow\cite{qu2025tokenflow} builds upon the features obtained from understanding- and generation-oriented tokenizers, and employs a Shared Mapping to project them into a Semantic Codebook and a Pixel Codebook, respectively. While this design facilitates cross-modal alignment, the shared mapping may not yield the optimal correspondence for either semantic abstraction or texture fidelity.
DualToken\cite{song2025dualtoken} introduces a hierarchical design within a single model: shallow layers are responsible for predicting pixel tokens, while deeper layers predict semantic tokens. During interaction with the LLM, the two token types are concatenated along the embedding dimension; during decoding, a Pixel Head and a Semantic Head are used for generation and understanding, respectively.
Finally, AToken\cite{lu2025atoken} identifies that many previous methods suffer from architectural inconsistency and modality-specific limitations. It proposes a fully Transformer-based unified tokenizer applicable to images, videos, and 3D scenes. By leveraging 4D Rotary Position Embedding (4D RoPE), AToken achieves both semantic understanding and high-fidelity reconstruction within a shared latent space, marking a key step toward true multimodal unification.

\subsubsection{Toward the Ideal Unified Tokenizer}
An ideal unified visual tokenizer should achieve an intrinsic unification of understanding and generation within a shared latent space, balancing semantic alignment with high-fidelity reconstruction.
Rather than maintaining the dichotomy between semantic-oriented and reconstruction-oriented tokenizers, it should encode visual content into representations that are simultaneously interpretable to language models and reversible to pixel-level details.




From a system perspective, such a tokenizer is expected to simultaneously satisfy several key properties: it must enable \textbf{understanding--generation compatibility}, supporting both high-level semantic reasoning and precise visual synthesis within a shared representational space; it should be \textbf{modality-extensible}, featuring a modular and flexible architecture that can be seamlessly extended to new modalities such as video, 3D, audio, or action; it needs to ensure \textbf{semantic consistency and reversibility}, so that encoding and decoding preserve stable semantic correspondences and maintain alignment between understanding and generation; and it must achieve \textbf{efficiency and compactness}, avoiding the redundancy and latency inherent in dual-branch designs through compact token representations and shared computation. In essence, an ideal unified tokenizer is not a simple fusion of understanding and generation, but a \textbf{semantically reversible, modality-agnostic, and structurally efficient} representation mechanism that lays the foundation for truly unified multimodal intelligence.






\section{Bridging Visual Coding and Visual Tokens: A Unified Perspective}

In this section, we bridge the gap between classical visual coding techniques and the visual token mechanisms employed in Multimodal Large Language Models (MLLMs). Despite their disparate origins---visual coding rooted in signal processing and compression standards, and MLLM tokens emerging from generative AI architectures---both paradigms share the common goal of representing visual information efficiently while preserving essential fidelity. We unify their principles through four key aspects: information theory, functionality, optimization, and objectives. This unification not only highlights intrinsic connections but also paves the way for cross-domain innovations in token technology.

\begin{figure}
    \centering
    \includegraphics[width=1\linewidth]{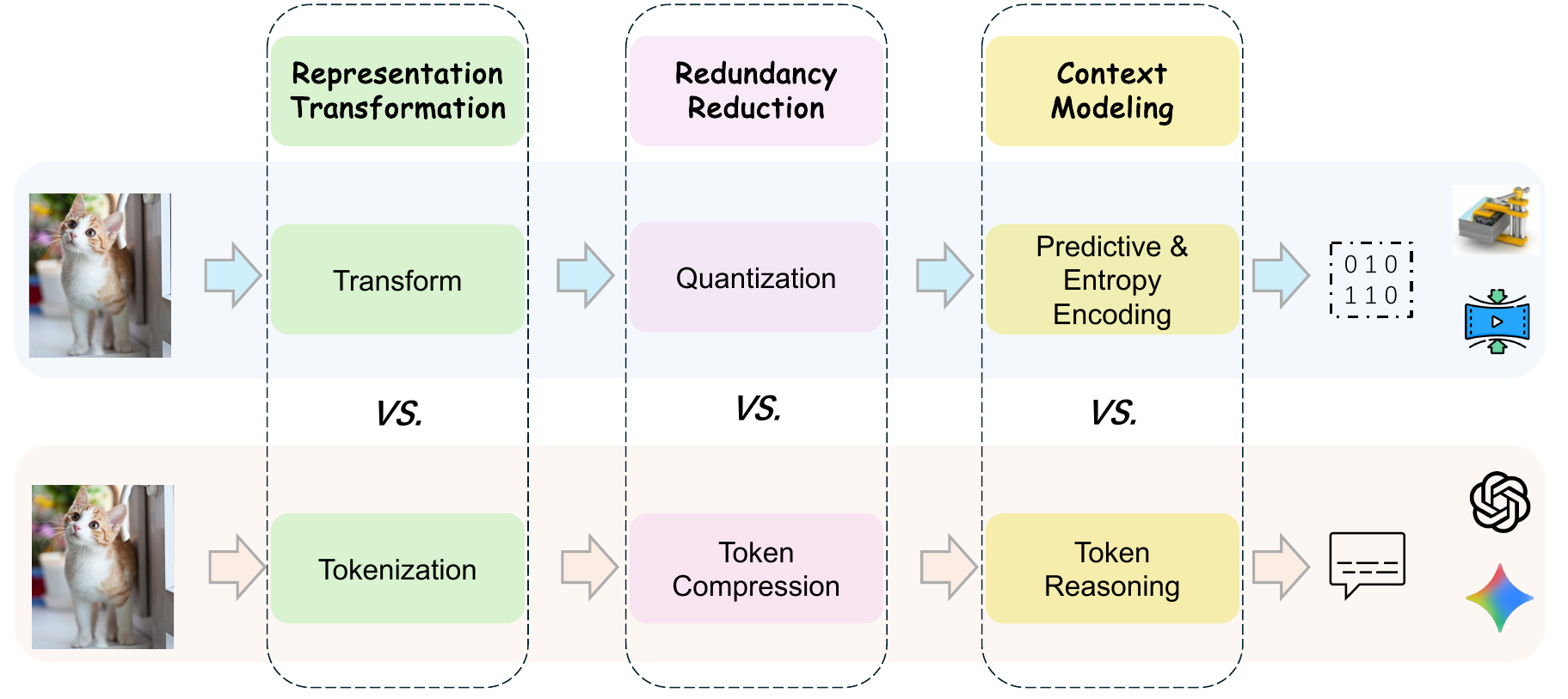}
    \caption{Analogy and comparison between Classical Visual Coding (top) and Visual Token Technology of MLLMs (bottom). Despite distinct operational modules of these two technologies, \textbf{both paradigms align under a shared functional workflow:} transforming raw inputs into latent representations (\textit{Representation Transformation}), compressing information by discarding non-essential details (\textit{Redundancy Reduction}), and capturing dependencies for downstream reconstruction or reasoning (\textit{Context Modeling}).}
    \vspace{-6mm}
    \label{fig:coding_vs_token}
\end{figure}

To intuitively illustrate this connection, Fig.~\ref{fig:coding_vs_token} presents a parallel view of the two paradigms. We map the distinct modules of classical coding (Transform, Quantization, Predictive \& Entropy Encoding) and MLLM processing (Tokenization, Token Compression, Token Reasoning) onto three shared functional stages: \textit{Representation Transformation}, \textit{Redundancy Reduction}, and \textit{Context Modeling}. 
In the classical view (top), raw pixels are transformed and quantized to remove statistical redundancy, resulting in a compact bitstream. Similarly, in the MLLM view (bottom), visual patches are tokenized and compressed to filter out semantic redundancy, forming a sequence of tokens ready for reasoning. This visual juxtaposition underscores that while the outputs differ—bitstreams for human perception versus semantic vectors for machine reasoning—the underlying structural logic remains remarkably consistent.

\subsection{Unified Formulation}

Here, we first try to understand \emph{Visual Token Technology} within the Multi-Modal Large Language Model (MLLM) pipeline~\cite{liu2023visual,zhu2024minigpt,bai2025qwen2} through the lens of the Information Bottleneck (IB) principle~\cite{goldfeld2020information,balle2018variational,minnen2018joint}. The core physical idea is that an optimal visual tokenizer must act as a strategic compressor similar to \emph{Visual Coding}, discarding irrelevant pixel-level details while zealously preserving the semantic information essential for downstream tasks like visual question answering. 

Let $X$ denote the raw visual input and $Z$ represent the compressed visual tokens. The original information bottleneck objective can be formulated as:
\vspace{-2.5pt}
\begin{equation}
\min_{p(z|x)} I(X;Z) - \beta I(Z;Y),
\end{equation}
\vspace{-2.5pt}
where $I(\cdot;\cdot)$ denotes mutual information, $Y$ is the target task, and $\beta$ controls the trade-off between compression and preservation. The fundamental IB objective, $\min I(X;Z) - \beta I(Z;Y)$, formalizes this trade-off: the first term $I(X;Z)$ represents the compression rate, penalizing the number of bits used to describe $X$ via $Z$, while the second term $I(Z;Y)$ is the relevance, rewarding the preservation of information about $Y$. The Lagrange multiplier $\beta$ controls the balance between these two competing goals; a high $\beta$ favors more descriptive but less compressed tokens. 

Based on the classic theory above, we can reform the process of Visual Tokenization as a compression problem. Specifically, the raw visual input $X$ is information-rich but highly redundant and often contains nuisances like texture and lighting variations. The goal is to find a compressed representation $Z$ (the visual tokens) that is maximally informative about the target task $Y$. So, we view the visual tokenization process $f_{\theta}: X \rightarrow Z$ as an information bottleneck:
\vspace{-2.5pt}
\begin{equation}
Z = f_{\theta}(X) = \arg\min_{Z} \underbrace{\mathcal{L}_{\text{comp}}}_{\text{Compression}} + \lambda \underbrace{\mathcal{L}_{\text{task}}}_{\text{Task Preservation}}
\end{equation}
\vspace{-2.5pt}
Specifically:
\vspace{-2.5pt}
\begin{align}
\mathcal{L}_{\text{comp}} &= \mathbb{E}_{x \sim p_{\text{data}}}[\|x - g_{\phi}(f_{\theta}(x))\|^2], \\
\mathcal{L}_{\text{task}} &= -\mathbb{E}_{(x,y) \sim p_{\text{data}}}[\log p_{\psi}(y|f_{\theta}(x))],
\end{align}
\vspace{-2.5pt}
where $g_{\phi}$ is a reconstruction decoder and $p_{\psi}$ is the task predictor. The visual tokenization function $f_\theta$ is thus conceptualized as an optimizer of this objective, practically achieved by minimizing a combination of a compression loss $\mathcal{L}_{\text{comp}}$ and a task preservation loss $\mathcal{L}_{\text{task}}$. The compression loss, often realized as a reconstruction error, ensures the tokens do not stray too far from the input data manifold, while the task loss, typically a cross-entropy loss, forces the tokens to be discriminative for the ultimate goal.

In this way, the optimal token representation $Z^*$ needs satisfy:
\vspace{-2.5pt}
\begin{equation}
p^*(z|x) = \frac{p^*(z)}{K(x,\beta)} \exp\left(-\beta \mathbb{E}_{y \sim p(y|x)}[D_{\text{KL}}(p(y|x) \| p(y|z))]\right),
\end{equation}
\vspace{-2.5pt}
where $K(x,\beta)$ is the normalization partition function. $p^*(z|x)$ reveals that the probability of a token $z$ given an input $x$ is proportional to its prior probability $p^*(z)$ re-weighted by an exponential factor of how well the token-induced conditional distribution $p(y|z)$ matches the true data distribution $p(y|x)$, with $\beta$ acting as the tuning parameter for this matching fidelity.

Moreover, to quantify the effectiveness of the tokenizer, the token efficiency ratio $\eta_{\text{token}}$ can be defined as the amount of task-relevant information per bit of compression, providing a single metric to evaluate different tokenization schemes:
\vspace{-2.5pt}
\begin{equation}
\eta_{\text{token}} = \frac{I(Z;Y)}{I(X;Z)} = \frac{\mathbb{E}_{z,y}[\log \frac{p(y|z)}{p(y)}]}{\mathbb{E}_{x,z}[\log \frac{p(z|x)}{p(z)}]},
\end{equation}
\vspace{-2.5pt}
which connects directly to the classical rate-distortion (R-D) theory, where the function $R(D)$ defines the fundamental limit of compression (the rate $R$) for a given maximum allowable distortion $D$ in reconstructing the input or its semantics.

Thus, for visual tokenizers, the rate-distortion trade-off follows:
\vspace{-5pt}
\begin{equation}
R(D) = \min_{p(z|x): \mathbb{E}[\Delta(X,g(Z))] \leq D} I(X;Z),
\end{equation}
where $\Delta$ is the distortion measure and $D$ is the maximum allowable distortion.

For hierarchical tokenization with scales $s_1 < s_2 < \cdots < s_k$, the information preservation is decomposed across hierarchical scales; the mutual information $I(X;Y)$ is approximated by a weighted sum of the information at each scale $I(Z_{s_i}; Y)$ minus the redundant information shared between consecutive scales $I(Z_{s_i}; Z_{s_{i+1}})$, ensuring that each level of the hierarchy captures unique and complementary semantic information, thereby achieving a more efficient and powerful visual representation for the MLLM:
\vspace{-2.5pt}
\begin{equation}
I(X;Y) \approx \sum_{i=1}^{k} \alpha_i I(Z_{s_i}; Y) - \sum_{i=1}^{k-1} \beta_i I(Z_{s_i}; Z_{s_{i+1}}),
\end{equation}
\vspace{-2.5pt}
where $\alpha_i$ and $\beta_i$ control information flow between scales.

In this way, we understand and formulate the popular visual token technology from the IB aspect, revealing its nature close to visual coding that pursues an information trade-off. Based on it, in the following section, we discuss visual coding and visual token technology in details from more specific perspectives, respectively.

\subsection{Information Theory Aspect: Shannon Entropy vs. Semantic Entropy}

\subsubsection{Unified Perspective}
From an information theory standpoint, both visual coding and MLLM tokenization can be seen as processes of entropy minimization. The core difference lies in the level of abstraction at which entropy is measured. Classical coding operates on the statistical properties of the signal (Shannon Entropy~\cite{balle2018variational,liu2023learned}), while MLLM tokenization operates on the meaning or conceptual information conveyed by the signal (Semantic Entropy~\cite{kuhn2023semantic,shani2025tokens}). In essence, tokenization is a semantic extension of classical coding, shifting the focus from compressing pixels to compressing concepts.

\subsubsection{Visual Coding: Minimizing Shannon Entropy}
In classical visual coding, the primary goal is to represent raw pixel data with the fewest bits possible. This is fundamentally governed by \textbf{Shannon entropy}, $H(X)$, which quantifies the theoretical lower bound for lossless compression based on the statistical redundancy of the source signal $X$. The objective is to design encoders that approach this limit by removing statistical correlations, thereby minimizing the bit-rate required for transmission or storage.
\vspace{-0.4em}
\subsubsection{MLLM Tokens: Minimizing Semantic Entropy}
MLLM tokenization is concerned with preserving \emph{meaning}, not exact pixel values. This motivates the adoption of \textbf{semantic entropy~\cite{kuhn2023semantic,shani2025tokens}}, $H_s(\tilde{U})$, which measures the uncertainty over a set of semantic equivalence classes. By collapsing signals that are syntactically different but semantically identical (e.g., two different images of a "cat on a mat"), semantic entropy is inherently lower than Shannon entropy ($H_s \le H$). MLLM encoders act as \emph{semantic filters}, discarding high-entropy, pixel-level details while retaining low-entropy, semantically decisive features. Thus, tokenization can be interpreted as compression guided not by Shannon entropy but by semantic entropy.
\vspace{-0.3em}
\subsection{Functionality Aspect: Redundancy Reduction vs. Context Modeling}

\subsubsection{Unified Perspective}
Functionally, both approaches achieve compression by building a probabilistic model of the visual data to exploit its structure. Classical coding explicitly models and removes statistical redundancy through fixed transforms and entropy coders. MLLM tokenization implicitly learns to model the deep semantic context and dependencies within the data through learned, high-capacity architectures like the transformer, which performs a sophisticated form of context-aware redundancy removal.
\vspace{-0.3em}
\subsubsection{Visual Coding: Redundancy Reduction}
Classical visual coding relies on explicit techniques for \textbf{redundancy reduction}. This typically involves a pipeline of decorrelation (e.g., Discrete Cosine Transform in JPEG~\cite{wallace1991jpeg}), which reduces spatial redundancy; quantization, which discards perceptually insignificant information; and entropy coding (e.g., Huffman or Arithmetic coding), which assigns shorter codes to more probable symbols. The functionality is to systematically strip away statistical redundancies present at the pixel level.

\subsubsection{MLLM Tokens: Context Modeling}
MLLM tokenization achieves compression through powerful \textbf{context modeling}~\cite{shani2025tokens,huang2024compression}. Vision transformers learn to represent an image as a sequence of tokens and use self-attention mechanisms to model the complex interdependencies between them. This process is analogous to next-token prediction in language models, where the model learns the conditional probability $P(\text{token}_t | \text{token}_{<t})$. By capturing the high-level semantic context, the model can form a compact representation that implicitly discards irrelevant details, effectively performing semantic compression.

\subsection{Optimization Aspect: R-D Trade-off vs. Information Bottleneck}

\subsubsection{Unified Perspective}
At their core, both domains solve an optimization problem that balances the compactness of the representation (rate) with its faithfulness to the original source (distortion). This can be universally formulated using the rate-distortion Lagrangian $\mathcal{L} = R + \lambda D$, where $\lambda$ controls the trade-off. The key distinction arises from how "rate" and "distortion" are defined. Moreover, these two problems are already tightly connected in information theory: the information bottleneck problem has a solution~\cite{goldfeld2020information} that exactly coincides with the single-letter rate–distortion formula for the remote source coding problem~\cite{dobrushin1962information, wolf1970transmission} under a logarithmic distortion function~\cite{courtade2013multiterminal}.

\subsubsection{Visual Coding: Rate-Distortion (R-D) Trade-off}
The classic \textbf{Rate-Distortion (R-D) trade-off} in visual coding is defined as:
\vspace{-2.5pt}
\begin{equation}
R(D) = \min I(X;Y) \quad \text{s.t.} \quad \mathbb{E}[d(X,Y)] \leq D,
\end{equation}
\vspace{-2.5pt}
where the rate ($R$) is measured in bits, and the distortion ($D$) is measured by perceptual metrics like Mean Squared Error (MSE) or SSIM. The goal is to find an encoding that uses the minimum number of bits for a given level of visual fidelity.

\subsubsection{MLLM Tokens: Information Bottleneck}
MLLM tokenization can be framed as an \textbf{Information Bottleneck} problem, which is a form of semantic rate-distortion optimization:
\vspace{-2.5pt}
\begin{equation}
R_s(D_s) = \min I(\tilde{X};\tilde{Y}) \quad \text{s.t.} \quad \mathbb{E}[d_s(\tilde{X},\tilde{Y})] \leq D_s,
\end{equation}
\vspace{-2.5pt}
Here, the "rate" ($R_s$) is operationalized by the number of tokens ($N$) or the computational complexity they induce (e.g., $\mathcal{O}(N^2)$), as this directly relates to the semantic code length and computational cost. The "distortion" ($D_s$) is semantic, measured by task performance metrics like classification accuracy or caption quality. The optimization seeks the most compact set of tokens that preserves the necessary semantic information for downstream tasks.

\subsection{Objective Aspect: Human Eye Fidelity vs. Machine Task Analysis}

\subsubsection{Unified Perspective}
The ultimate objective of any compression scheme is to preserve the fidelity of the information for its intended "user." Both visual coding and MLLM tokenization are optimized for a specific user, but the nature of this user differs fundamentally. This leads to distinct definitions of what constitutes acceptable information loss.

\subsubsection{Visual Coding: Human Eye Fidelity}
Classical visual coding is designed for human consumption. Therefore, its primary objective is to maintain high \textbf{human eye fidelity}. The distortion metrics (e.g., PSNR, SSIM, VMAF) are engineered to correlate with the human visual system's perception of quality. The goal is to create a compressed representation that is perceptually indistinguishable, or nearly so, from the original to a human observer.

\subsubsection{MLLM Tokens: Machine Task Analysis}
In contrast, MLLM tokens are generated for machine consumption. The objective is not perceptual quality but successful \textbf{machine task analysis}. The fidelity of the tokenized representation is measured by its utility in downstream tasks, such as image classification, object detection, or visual question answering. Therefore, the system is optimized to preserve task-relevant semantic features, even if this comes at the cost of pixel-level accuracy that would be noticeable to a human. This bridging framework sets the stage for subsequent discussions on multimodal tokens and their applications in communication and embodied AI.

\begin{table*}[t]
\centering
\caption{Comparison between classical visual coding and MLLM tokenization under the unified framework.}
\vspace{-8pt}
\begin{tabular}{p{3.5cm}p{5.5cm}p{5.5cm}}
\toprule
\textbf{Aspect} & \textbf{Classical Visual Coding} & \textbf{MLLM Tokenization} \\
\midrule
\textbf{1. Information Theory} 
& Minimize \textbf{Shannon Entropy} (statistical uncertainty)
& Minimize \textbf{Semantic Entropy} (conceptual uncertainty) \\
\midrule
\textbf{2. Functionality} 
& Explicit \textbf{Redundancy Reduction} (e.g., DCT, entropy coding)
& Learned \textbf{Context Modeling} (e.g., self-attention in transformers) \\
\midrule
\textbf{3. Optimization} 
& \textbf{Rate-Distortion (R-D) Trade-off} (Rate in bits, Distortion in perceptual error)
& \textbf{Information Bottleneck} (Rate in tokens/compute, Distortion in task error) \\
\midrule
\textbf{4. Objective} 
& Preserve \textbf{Human Eye Fidelity} (for human viewers)
& Enable \textbf{Machine Task Analysis} (for machine algorithms) \\
\bottomrule
\end{tabular}
\vspace{-16pt}
\end{table*}

\subsection{How Visual Coding Principles Can Refine Token Technology}

The maturity of classical visual coding provides a rich set of optimization tools that can be directly adapted to mitigate inefficiencies in current visual tokenizers. By casting token generation as signal compression, we can inject structural priors and principled rate control into the tokenization pipeline.

\textbf{Structural Decorrelation and Transformation} 
Current tokenizers often treat image patches as independent units or rely solely on self-attention to find correlations. Classical coding suggests that transforming signals into a decorrelated domain significantly enhances compressibility. Inspired by this, recent works have explored operating in the frequency domain via discrete transforms to compact energy before tokenization~\cite{feng2024docpedia,pertsch2025fast}. Furthermore, borrowing the concept of Inter/Intra-frame coding from video standards, temporal redundancy can be explicitly modeled. For instance, logic similar to Group of Pictures (GOP) structures can be applied to token streams, separating information-rich ``key-tokens'' from predictable ``motion-tokens,'' thereby initializing a far more efficient representation for video inputs~\cite{gadot2025rl}.

\textbf{Entropy-Aware Token Management} 
Standard ViTs produce a fixed number of tokens regardless of content complexity, a stark contrast to the variable-bitrate nature of efficient codecs. Principles from entropy coding, such as Run-Length Encoding (RLE), serve as a blueprint for merging consecutive, redundant tokens in semantic space~\cite{choudhury2024don}. Moving beyond simple heuristics, the rigorous rate-control philosophy—optimizing the trade-off between bit consumption and distortion—can be adapted into ``information-preserving guided selection.'' This involves pruning or retaining tokens based on their marginal contribution to the total semantic information, effectively applying rate-distortion optimization (RDO) to the token budget~\cite{tan2025tokencarve}.

\textbf{Complexity-Adaptive Representation} 
Underpinning optimal compression is the principle of Minimum Description Length (MDL), a computable proxy for Kolmogorov Complexity. Applying this to MLLMs advocates for variable-length tokenization mechanisms~\cite{duggal2025single}. Instead of a uniform grid, the tokenizer should dynamically allocate fewer symbols to simple, low-frequency regions and more symbols to complex, high-frequency details. This mirrors the quantization parameter (QP) adaptation in codecs, ensuring that the token count scales linearly with the semantic density of the input.

\textbf{Discretization via Vector Quantization} 
While continuous embeddings dominate understanding tasks, the stability of storage and transmission benefits from the discrete nature of digital signals. Vector Quantization (VQ) acts as the bridge, mapping continuous latent spaces to discrete codebooks. This process not only aligns with the symbolic nature of language models but has been proven to stabilize generative tasks and reduce representation costs by enforcing a compact, learned vocabulary~\cite{geng2025x}.



\begin{figure}[!t]
\centering
\includegraphics[width=1\linewidth]{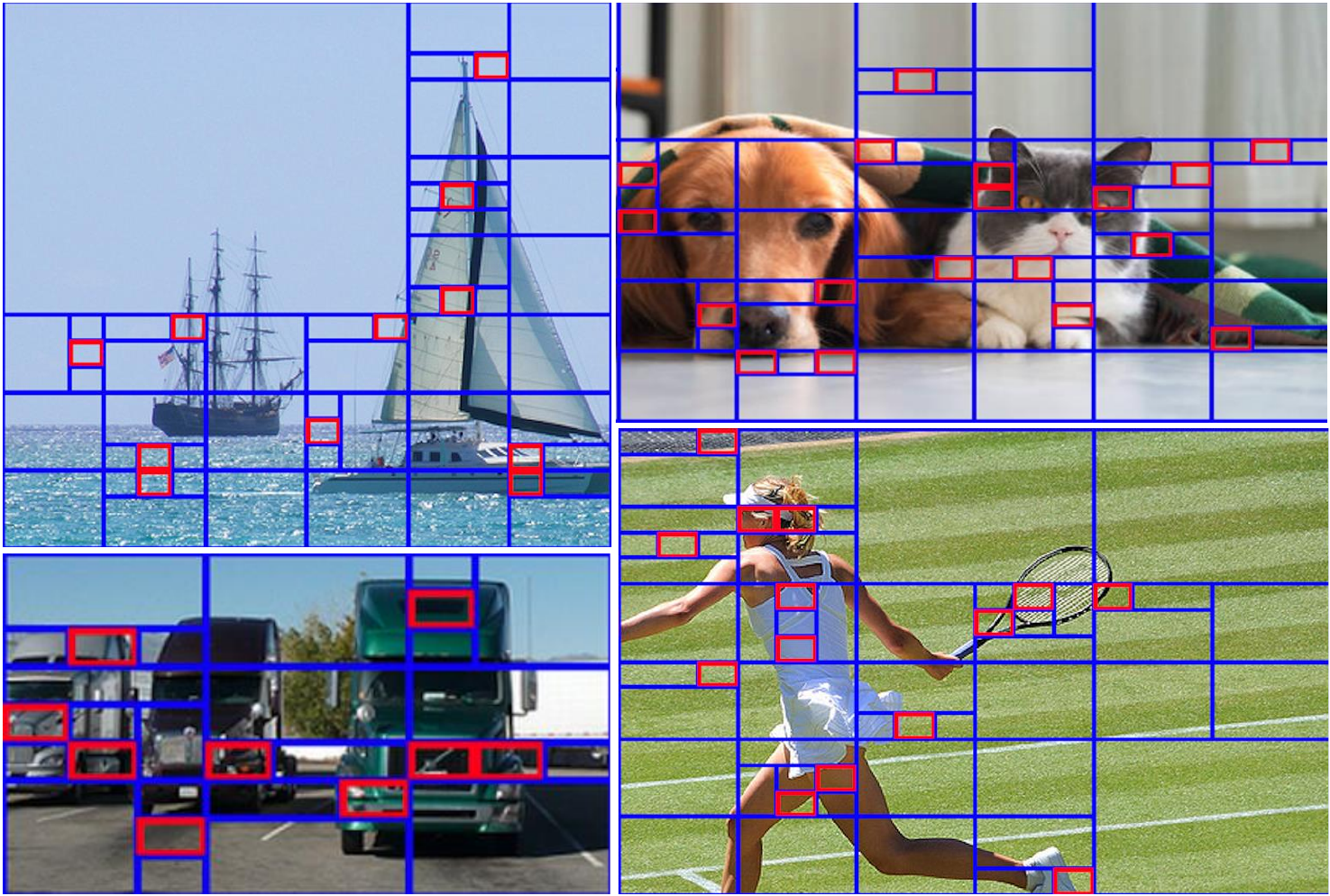}
\vspace{-16pt}
\caption{Adaptive quadtree partitioning driven by information density. Low-information regions remain coarse; high-information regions are split more finely. Red boxes denote retained dense tokens.}
\label{quadtree}
\vspace{-16pt}
\end{figure}

\begin{table}[!t]
\centering
\footnotesize
\caption{\textbf{LLaVA-v1.5~\cite{liu2023visual} (7B) under three visual-token budgets.} We compare QPID with FastV~\cite{chen2024image} and PruMerge~\cite{shang2024prumerge} on six benchmarks at 25\%/12.5\%/6.25\% retention (144/72/36 tokens). ``Vanilla'' uses all 576 tokens. The last column reports accuracy as a percentage of the full-token average; best in \textbf{bold}.}
\label{tab:qpid_llava7b}
\setlength{\tabcolsep}{3.5pt}
\begin{tabular}{@{}l|cccccc|c@{}}
\toprule[1.5pt]
\multicolumn{8}{c}{\textbf{LLaVA-v1.5-7B Results}} \\
\midrule
\textbf{Method} & \textbf{MME} & \textbf{SciQA} & \textbf{VQA\textsuperscript{T}} & \textbf{POPE} & \textbf{SeedB} & \textbf{VizW} & \textbf{Avg} \\
\midrule
\rowcolor{gray!15}
Vanilla & 1512.53 & 53.76 & 57.62 & 85.5 & 65.79 & 53.21 & 100.0\% \\
\midrule
\multicolumn{8}{c}{\textit{Ratio = 25\% (144 tokens)}} \\
\midrule
FastV      & 1396.66 & 52.79 & 51.10 & 73.7 & 61.94 & 50.93 & 92.55\% \\
PruMerge   & \textbf{1416.52} & 53.53 & 55.03 & 81.6 & \textbf{62.39} & 52.04 & 96.13\% \\
\textbf{QPID} & 1415.67 & \textbf{54.60} & \textbf{55.24} & \textbf{85.8} & 62.19 & \textbf{52.33} & \textbf{96.82}\% \\
\midrule
\multicolumn{8}{c}{\textit{Ratio = 12.5\% (72 tokens)}} \\
\midrule
FastV      & 1301.68 & 52.96 & 48.65 & 62.7 & 56.40 & 50.97 & 85.51\% \\
PruMerge   & 1348.89 & 53.05 & 54.12 & 74.7 & 58.48 & 53.00 & 91.72\% \\
\textbf{QPID} & \textbf{1353.52} & \textbf{53.59} & \textbf{54.34} & \textbf{84.3} & \textbf{59.76} & \textbf{53.42} & \textbf{94.17}\% \\
\midrule
\multicolumn{8}{c}{\textit{Ratio = 6.25\% (36 tokens)}} \\
\midrule
FastV      & 1054.14 & 51.73 & 45.97 & 43.3 & 49.03 & 50.07 & 74.64\% \\
PruMerge   & 1261.41 & 52.86 & 52.98 & 67.5 & 54.09 & 52.61 & 86.64\% \\
\textbf{QPID} & \textbf{1308.56} & \textbf{52.89} & \textbf{53.01} & \textbf{77.5} & \textbf{56.32} & \textbf{53.81} & \textbf{90.22}\% \\
\bottomrule[1.5pt]
\end{tabular}
\vspace{-18pt}
\end{table}


\begin{figure}[!t]
\centering
\includegraphics[width=1\linewidth]{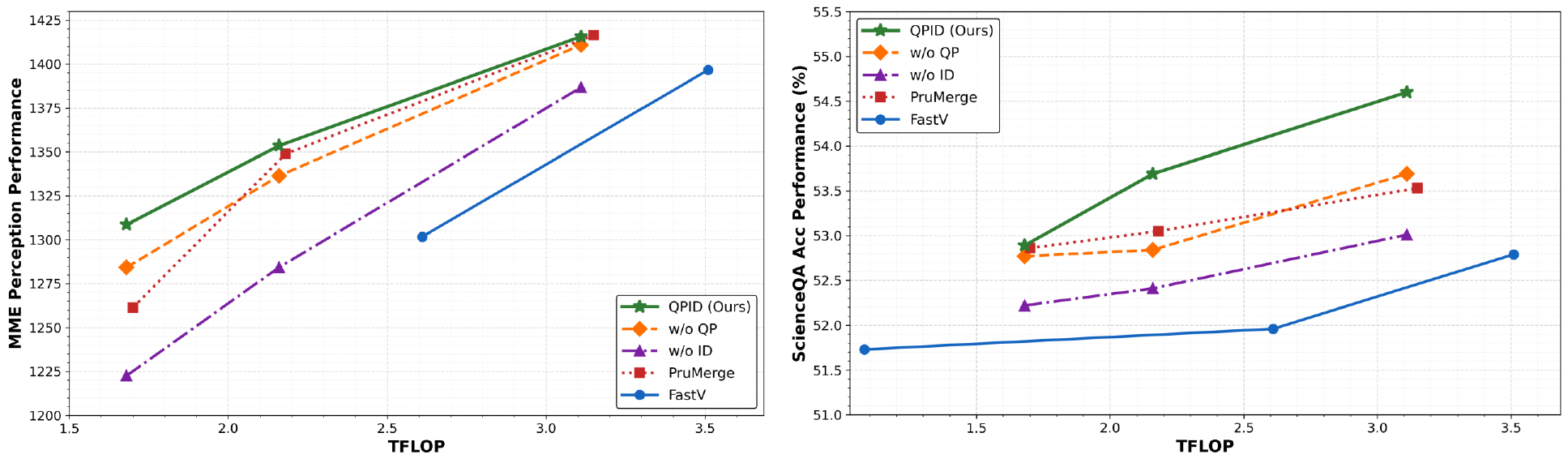}
\vspace{-18pt}
\caption{Ablation of visual‐token pruning on LLaVA-v1.5-7B~\cite{liu2023visual}. We plot MME-Perception~\cite{Fu2023MMEAC} (left) and ScienceQA~\cite{saikh2022scienceqa} accuracy (right) vs.\ actual TFLOPs under three token‐retention rates (25\%, 12.5\%, 6.25\%). QPID (ID+QP) consistently leads across all budgets.}
\label{ablation}
\vspace{-18pt}
\end{figure}


\textbf{Case Study I: Quadtree Partitioning–based Visual Token Pruning Considering Information Density.} We replace attention score heuristics with an information-theoretic, structure-aware pipeline. First, an \emph{entropy-based information density} criterion selects a compact, non–redundant subset of visual tokens, distributing them across the scene rather than clustering in a few high–attention regions; this task–agnostic scoring suppresses background redundancy and preserves diverse cues. Second, an adaptive \emph{quadtree partitioning} allocator refines spatial granularity where information is high while keeping homogeneous areas as large leaves, so more tokens are assigned to semantically rich zones without losing global layout. The four–panel visualization in Fig.~\ref{quadtree} illustrates this behavior: low information regions remain coarse, high information regions are split more finely, and red boxes mark retained dense tokens. Quantitatively, Table~\ref{tab:qpid_llava7b} (LLaVA-v1.5-7B~\cite{liu2023visual}) shows that \textbf{QPID} attains the best overall accuracy at all three budgets: at \textbf{25\%} tokens it maintains \textbf{96.82\%} of the full-token average and leads on four of six benchmarks; at \textbf{12.5\%} it reaches \textbf{94.17\%} and is best on all tasks; and at the extreme \textbf{6.25\%} (36 tokens) it still preserves \textbf{90.22\%}, widening the margin over prior methods as the budget tightens. The ablations in Fig.~\ref{ablation} further isolate contributions under matched compute: on both MME-Perception~\cite{Fu2023MMEAC} (left) and ScienceQA~\cite{saikh2022scienceqa} accuracy (right), the full \textbf{QPID} (ID{+}QP) curve consistently lies above its variants across the 25\%/12.5\%/6.25\% settings; removing information-density scoring (WO/ID) produces the largest drop, while omitting quadtree partitioning (WO/QP) also degrades results. Together, these findings indicate that entropy-driven selection and adaptive quadtree allocation are jointly responsible for stable accuracy at very small token budgets and deliver favorable accuracy–compute trade-offs for multimodal inference.

\subsection{How Token Technology Can Refresh Codecs}
Conversely, the rise of MLLMs and visual token Technology introduces a semantic dimension to the traditional signal processing field. The powerful reasoning and predictive capabilities of these models are transforming codecs from pixel-matching engines into semantic-aware intelligence systems.

\textbf{Semantic-Guided Rate Allocation.} 
Traditional codecs struggle to distinguish between statistically complex noise and semantically important details. MLLMs can serve as a perceptual ``brain'' for the codec, analyzing the scene to generate semantic importance maps. These maps guide the encoder to allocate high bitrates to critical regions—such as text or human faces—while aggressively compressing irrelevant backgrounds, thus optimizing the bitstream for downstream machine vision utility rather than purely human perceptual metrics~\cite{liu2024tell,li2024misc}.

\textbf{Feature-Domain Compression.} 
As the consumers of visual data shift from humans to machines, the optimal compression target shifts from pixels to intermediate representations. A new paradigm of ``Token Coding'' is emerging, where the bitstream directly encapsulates semantic tokens rather than reconstructed pixels. This approach is particularly valuable for edge-cloud systems; by placing the tokenizer at the edge, one can transmit compact feature tokens~\cite{gao2024feature,qiao2025token} or specialized machine-oriented bitstreams~\cite{kaobridging,li2024high}, significantly reducing bandwidth requirements while preserving the performance of cloud-based MLLMs.

\textbf{Universal Probabilistic Modeling.} 
At its core, compression is about prediction: the better one can predict the next symbol, the fewer bits are needed to encode it. MLLMs, trained on big data, have emerged as powerful general-purpose predictors. Their ability to model long-range dependencies and complex patterns allows them to function as universal compressors~\cite{li2025lossless,deletanglanguage}. By treating raw data bytes as tokens, MLLMs have demonstrated the potential to surpass specialized engineering codecs (like PNG) in compression ratios, hinting at a future where intelligence and compression are unified under a single probabilistic framework.

\begin{figure}
    \centering
    \includegraphics[width=1\linewidth]{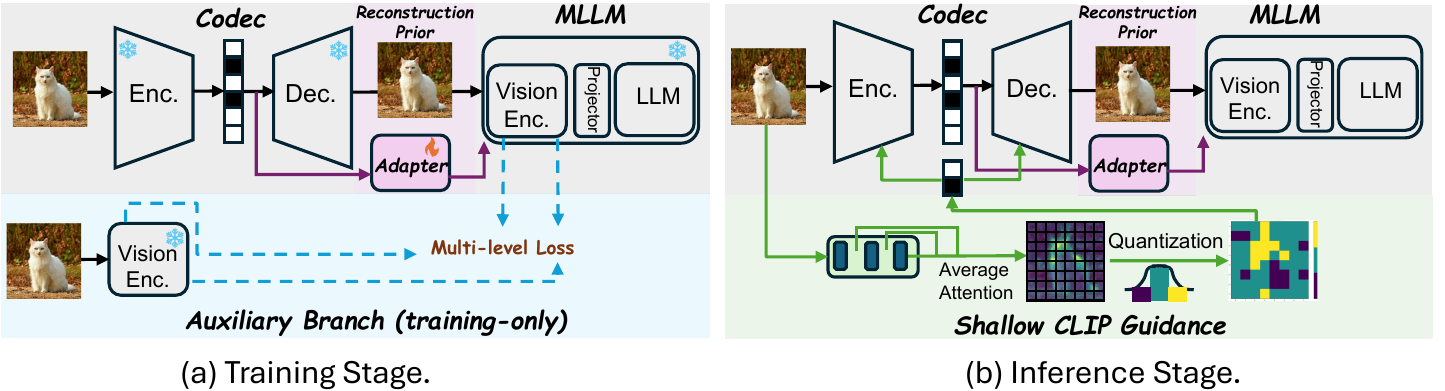}
    \caption{The framework of the Coding Paradigm Tailored to MLLMs (CoTAM)~\cite{liu2025mllms_cotam}. This paradigm utilizes the CLIP token-level prior to help improve the performance on compressed images.}
    \vspace{-6mm}
    \label{fig:case_II_framework}
\end{figure}

\begin{figure*}
    \centering
    \includegraphics[width=1\linewidth]{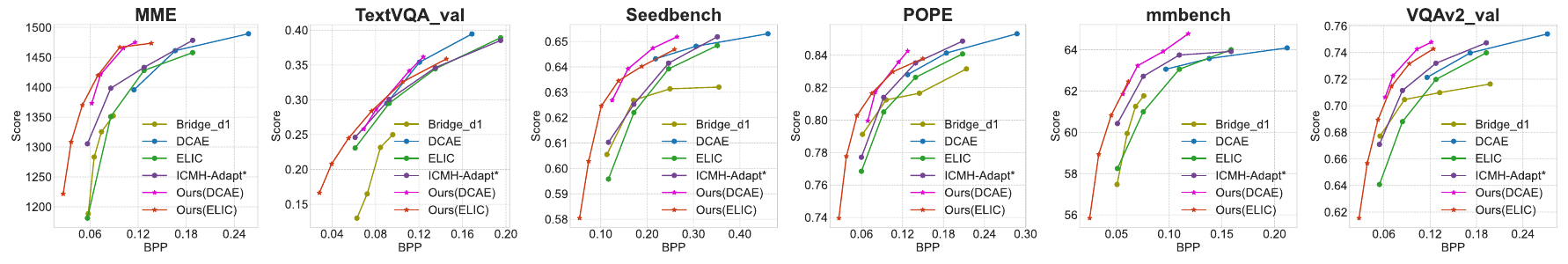}
    \vspace{-18pt}
    \caption{The results of CoTAM~\cite{liu2025mllms_cotam}. By utilizing token-level CLIP guidance, compared with recent codecs (ELIC~\cite{he2022elic}, DCAE~\cite{lu2025learned_dcae}, Bridge~\cite{kaobridging}, Adapt-ICMH~\cite{li2024image}), it achieves better performance on MLLM tasks.}
    \label{fig:case_II_results}
    \vspace{-18pt}
\end{figure*}

\textbf{Case Study II: A Coding Paradigm Tailored to MLLMs.} Recent research~\cite{liu2025mllms_cotam} challenges the traditional decoupling of coding and machine perception by proposing CoTAM, a codec explicitly tailored for MLLMs. Instead of treating the downstream model as a black box, this approach analyzes the internal information flow of the vision encoder (e.g., CLIP), identifying a distinct "three-stage" processing pattern: preliminary screening, local extraction, and global semantic integration. The study reveals a critical insight: compression distortion does not affect all tokens uniformly; it disproportionately disrupts the "cross-level" features where low-level structural details are synthesized into high-level semantics.
As shown in Fig.~\ref{fig:case_II_framework}, guided by this token-level intelligence, CoTAM introduces a Shallow CLIP-Guided mechanism. It extracts attention maps from the shallow layers of the vision encoder to generate a semantic importance map, which directly controls the quantization step in the image codec to allocate more bits to semantically rich regions. Furthermore, it employs a multi-level fidelity decoder to align the reconstructed signal with the MLLM's feature hierarchy. By achieving up to ~36\% bitrate savings on six benchmarks (MME~\cite{Fu2023MMEAC}, TextVQA~\cite{singh2019towards}, POPE~\cite{li2023evaluating}, SeedBench~\cite{li2023seed}, VQAv2~\cite{goyal2017making}, MMMU~\cite{yue2024mmmu}, and MMBench~\cite{liu2024mmbench}, as shown in Fig.~\ref{fig:case_II_results}) for comparable MLLM performance, CoTAM exemplifies the potential of "Compression Tells Intelligence": leveraging the model's own token attention mechanisms to optimize the fundamental rate-distortion trade-off in signal coding.

\section{Application and Outlook}




\subsection{Next-generation Token Applications}
\subsubsection{Token Technology in AIGC}
Token technology is increasingly central in the AIGC era, where models must map high-dimensional continuous signals to compact representations for efficient generation. Across modalities, tokenization is converging toward a shared goal: \emph{compact, semantically rich, and generative-friendly} representations. Text tokenization provides the discrete modeling blueprint\cite{kudo2018sentencepiece}, image tokenization extends it to perceptual semantics\cite{esser2021taming}, and video tokenization largely inherits and adapts image techniques\cite{yu2023magvit}.

\textbf{Image tokenization.} Image generation like controllable synthesis and editing demands representations that preserve semantics while remaining efficient. Autoregressive systems rely on discrete tokens as the interface between images and LLM modeling\cite{ramesh2021dalle}. Diffusion models originally introduced latent tokenization to reduce spatial resolution and accelerate training/inference\cite{rombach2022ldm}. As both paradigms scale, they increasingly converge: AR models are bottlenecked by long visual token sequences and thus seek compact-yet-detailed tokenizers\cite{chen2024image}, while diffusion models push toward more semantically aligned latent spaces to improve global coherence and controllability\cite{yu2024representation}. This convergence suggests tokenization is a key leverage point where coding principles and learned compression jointly shape modern AIGC.

\textbf{Video tokenization.} Current video generation pipelines often combine frame-wise tokenization with temporal attention and redundancy reduction\cite{kondratyuk2023videopoet,Magvit-v2}, but a principled temporal tokenizer remains less mature. A promising path is to re-introduce classical video-coding insights (motion prediction, hierarchical redundancy removal) into learned token pipelines\cite{liu2025revisiting}.

In summary, tokenization across image, video is becoming increasingly unified around compact and semantically aligned representations\cite{lu2025atoken}, with image tokenization acting as the primary driver and video tokenization as a fast-growing frontier.

\subsubsection{Token Technology in Embodied AI}

Embodied AI increasingly uses end-to-end foundation models that unify perception, language, and control. Here, tokenization acts as a machine-native compression interface, converting high-dimensional sensory streams (and optionally actions) into compact sequences for LLM-style backbones, improving data efficiency for long-horizon reasoning and real-time control.

\textbf{Perception and context compression.}
For manipulation, VLA systems often use continuous visual tokens from ViT features (RT-2\cite{brohan2023rt2}, OpenVLA\cite{kim2024openvla}), with OpenVLA combining semantically aligned tokens via SigLIP\cite{zhai2023sigmoid} and geometry-rich tokens via DINOv2\cite{oquab2024dinov2}. Discrete tokenization via VQ-style codebooks enables scalable world modeling and next-token rollout (Genie\cite{bruce2024genie}), while object-centric sparsity reduces redundancy by encoding salient entities (VIMA\cite{jiang2023vima}). To handle long temporal contexts under transformer complexity, systems apply dynamic token pruning for real-time efficiency (LightVLA\cite{jiang2025lightVLA}, FAST\cite{pertsch2025fast}) and compress historical interactions into highly compact memory tokens for retrieval (MemoryVLA\cite{shi2025memoryvla}).

\textbf{Action tokenization.}
Recent models also compress continuous control into discrete action tokens, aligning robot outputs with the sequence modeling interface. RT-2\cite{brohan2023rt2} uses explicit quantization, while learned codebooks mitigate multi-modal ``average action'' effects (VQ-BeT\cite{lee2024VQ-BeT} built on VQ-VAE\cite{oord2017vqvae}), enabling action generation as discrete token prediction.

\subsubsection{Categorization and Transferability of Visual Tokenizers}
Building upon the taxonomy in Section~\ref{sec:unified_tokenizer}, we revisit tokenizer families from the lens of transferability. Understanding-oriented tokenizers (CLIP-ViT\cite{radford2021clip}, SigLIP\cite{zhai2023sigmoid}, Perception Encoder\cite{bolya2025perception}) emphasize semantic abstraction and cross-modal alignment, making them reusable across architectures with lightweight adaptation. Generation-oriented tokenizers (VQ-VAE\cite{oord2017vqvae}, VQGAN\cite{esser2021taming}) prioritize high-fidelity reconstruction, but heterogeneous codebooks and objectives can hinder portability. Unified tokenizers (e.g., Show-o2\cite{xie2025show}) attempt to encode semantics and details in one space, yet principled mechanisms to balance objectives and ensure cross-model portability remain underexplored. Overall, understanding-oriented tokenizers are typically more transferable, while generation-oriented tokenizers offer stronger perceptual fidelity but face practical transfer challenges.

\subsection{Next-generation Codec Applications}
\subsubsection{Immersive Media}
NeRF-based codecs.  Neural radiance fields (NeRF) have rapidly evolved from pure view-synthesis models into neural codecs for static and dynamic 3D content. Instead of transmitting per-frame pixels, NeRF-based methods encode a compact radiance field whose parameters are optimized for novel-view rendering, and then quantize and entropy-code these parameters as the bitstream. For static scenes, NeRFCodec \cite{li2024nerfcodec} is a representative end-to-end design: it treats NeRF feature planes as latent images, reuses a pretrained 2D neural image codec, and learns lightweight scene-specific encoder/decoder heads under a joint rendering and rate–distortion objective, thereby achieving high-quality novel-view synthesis from bitstreams on the order of a few hundred kilobytes. For dynamic content, several works explicitly cast NeRF as a volumetric video codec. VRVVC \cite{hu2025vrvvc} further introduces a tri-plane residual representation together with learnable quantization and compact entropy models, enabling variable-rate volumetric video compression with a single network and competitive rate–distortion performance across a wide bitrate range. Streaming radiance fields \cite{li2022streaming} demonstrate that explicit-grid radiance fields can also be updated over time and transmitted via model-difference coding, paving a path toward online NeRF-style streaming. Conceptually, these systems can be regarded as NeRF-based token codecs: structured radiance-field tokens (grid cells, tri-plane coefficients, residual fields, latent planes) become the basic symbols, and codec design focuses on their parameterization, on bit allocation between geometry and appearance, and on integration with conventional streaming infrastructures.

\subsubsection{MLLMs for Codec}
With multimodal LLMs as receivers, codec objectives increasingly shift from classical rate--distortion (RD)\cite{harell2025rate} toward \emph{rate--task performance} (RT): under a bit budget, maximize downstream utility while keeping latency and memory bounded. Two directions are emerging. (i) \emph{MLLM-aware codecs} optimize representations for machine receivers, including end-to-end task losses\cite{le2021image}, unified human/machine coding with multimodal supervision\cite{yin2025unified}, and compression tailored to VLM decoders with explicit RT trade-offs\cite{li2024high}. (ii) \emph{(M)LLMs as priors/decoders} leverage generative sequence models for compression: theory connects language modeling and compression\cite{deletang2023language}, and recent systems demonstrate LLM-assisted lossless image coding via visual prompting\cite{du2025large} or language-space prediction\cite{chen2024large}. These trends motivate evaluating codecs by RT curves (bitrate vs.\ task performance) and system metrics (decoding latency, KV-cache footprint), not RD alone.

\subsubsection{Video Coding for Machine (VCM)}
VCM targets scenarios where the consumer is a machine (detector/tracker/MLLM), and the bitstream may carry pixels, intermediate features, or semantic descriptors. MPEG exploratory work formalizes tracks, common test conditions, and evaluation protocols that distinguish signal-domain and feature-domain pipelines\cite{mpeg_explorations_34_vcm,mpeg2025vcmexplore,lee2023vcmstandard}. Representative directions include task-aware signal coding\cite{zhang2024smr}, intermediate feature compression with notable RD/complexity advantages\cite{kim2023e2evcm,lee2023transformvcm}, and semantic-level coding/collaborative analytics summarized in surveys\cite{yang2024vcm}. Emerging theory further studies RT limits and rate--accuracy bounds for analysis tasks\cite{liu2025rtbounds}, while standardization efforts continue to converge on test protocols and metrics\cite{itu2022vcm}.

\subsection{Unified Communication System in The LLM Era}
Modern intelligent receivers motivate a representation-centric view of communication, driven by \emph{what} is communicated and \emph{why}. We distinguish systems by (i) communication unit (bits/semantics/tokens), (ii) objective (signal fidelity/task utility/model-conditioned utility under compute/memory budgets), and (iii) receiver interface (reconstruction/semantic inference/foundation-model conditioning).

\subsubsection{Traditional Communication}
Classical communication optimizes bit recoverability: source coding removes redundancy toward entropy, channel coding protects bits under noise, and separation motivates independent design\cite{shannon1948mathematical,nalewajski2011elements}. Distortion is defined in signal space (e.g., MSE/PSNR)\cite{fardo2016formal}, and deep Joint Source-Channel Coding (JSCC) variants largely remain reconstruction-oriented\cite{bourtsoulatze2019deepjscc,kurka2021bandwidth,yang2021deep}. Finite-blocklength theory clarifies the gap to Shannon limits under short packets\cite{polyanskiy2010channel,mary2016finite,kostina2012fixed} and extends to one-shot and multiuser settings\cite{yassaee2013technique,li2021unified,scarlett2014second}, explaining why reconstruction-driven RD pipelines can mismatch modern machine receivers.

\subsubsection{Semantic Communication}
Semantic communication shifts to task utility by transmitting only what is needed for a task. Task-oriented JSCC directly optimizes downstream losses over noisy channels\cite{xie2021deepscjsac,bourtsoulatze2019deepjscc,xie2021taskvqa}. Surveys and tutorials formalize the rate--task viewpoint and evaluation beyond RD\cite{yang2022semantic,lu2023semantics,gunduz2024joint}. Foundation models (multimodal LLMs, diffusion priors) can serve as semantic front-ends for summarization and regeneration\cite{jiang2024large,wang2025large}, though generalization across tasks/channels and short-blocklength overheads remain challenges; robust task-oriented training is an active direction\cite{park2025robust}.

\subsubsection{Token Communication}
\noindent \textit{Machine-native coordination via learned tokens.}
Token communication advances semantic communication by adopting model-consumable tokens as the transmitted interface\cite{kang2017neurosurgeon,teerapittayanon2017distributed}. A sensor agent maps observations into compact continuous embeddings or discrete codebook indices for direct consumption, aligning with split inference and feature transmission to reduce latency and on-device compute\cite{eshratifar2019bottlenet}. Learned machine-language tokens can be substantially more transmission-efficient than natural language while remaining task-sufficient\cite{xiao2025machinetokens}.

Robustness can be addressed via \textbf{Joint Token \& Channel Coding (JTCC)} that injects channel impairments during training\cite{xiao2025machinetokens}, or via \textbf{analog mappings} that transmit token vectors directly (``over-the-air tokens'') leveraging deep JSCC and over-the-air computation\cite{kurka2021bandwidth,yang2021deep,goldenbaum2013harnessing,sahin2023oacsurvey,nazer2007computation}. Recovered tokens can condition an LLM through soft-prefix prompting without fine-tuning\cite{li2021prefix,lester2021power}. Recent token-domain multi-access designs further explore contextual prediction to mitigate collisions and improve bandwidth efficiency\cite{qiao2025todma,qiao2025tokcom}.

\section{Conclusion}
Guided by the principle that ``Compression Tells Intelligence,'' this paper unifies classical visual coding and emerging visual token technology under a shared view of efficiency--fidelity trade-offs. We connect the two through a common framework spanning information measures (Shannon vs.\ semantic), functional roles (redundancy reduction vs.\ context modeling), optimization criteria (R--D vs.\ information bottleneck), and objectives (human fidelity vs.\ machine utility), etc.
This unification yields bidirectional insights: coding principles (e.g., decorrelation and entropy-aware rate control) can improve token systems, while token-based semantic modeling motivates next-generation codecs optimized for machine tasks. We also discuss the potential impacts of the token techniques on MLLMs, AIGC, and even embodied AI, and outline the next generation of visual coding technology.
Future work includes unified tokenizers balancing semantic alignment and reconstructive fidelity, token communication across platforms, and extending the framework to emerging modalities such as 3D and 4D information.

\section*{Acknowledgments}
This work was supported in part by NSFC 62302246 and ZJNSFC under Grant LQ23F010008, and supported by High Performance Computing Center at Eastern Institute of Technology, Ningbo, and Ningbo Institute of Digital Twin.\looseness=-1

\scriptsize
\bibliographystyle{IEEEtran}
\bibliography{references}

\newpage

\end{document}